\pdfoutput=1
\documentclass[11pt]{article}

\usepackage[final]{acl}

\usepackage{times}
\usepackage{latexsym}

\usepackage[T1]{fontenc}
\usepackage[utf8]{inputenc}

\usepackage{amssymb}
\usepackage{xspace}
\usepackage{booktabs}
\usepackage[table]{xcolor}
\usepackage[most]{tcolorbox}

\tcbset{
  promptbox/.style={
    width=\linewidth,
    top=8pt,
    bottom=2pt,
    left=4pt,
    right=4pt,
    colback=gray!1!white,
    colframe=black,
    colbacktitle=black,
    enhanced,
    center,
    fontupper=\normalsize,
    before upper={\setlength{\baselineskip}{1.12\baselineskip}}, % <- 加这一行
    attach boxed title to top left={yshift=-0.1in,xshift=0.15in},
    boxed title style={boxrule=0pt,colframe=white,},
  }
}

\newtcolorbox{promptbox}[2][]{promptbox,title=#2,#1}
\newcounter{case}
\newtcolorbox{casebox}[2][]{%
  before upper={\refstepcounter{case}},
  colframe=black!80!black,
  colback=gray!10!white,
  colbacktitle=black!80!white,
  coltitle=white,
  boxrule=0.5mm,
  enhanced,
  rounded corners,
  width=\linewidth,
  title=\textbf{Case Study #2},
  #1
}

\usepackage{capt-of}
\usepackage{tabularx}
\usepackage{array}
\definecolor{mygreen}{HTML}{298D66}
\definecolor{myred}{HTML}{F94153}
\definecolor{tts_or}{HTML}{8AD879}
\definecolor{tts_sc}{HTML}{53D2DC}
\definecolor{grpo}{HTML}{3C5488}
\definecolor{gspo}{HTML}{8491B4}
\definecolor{srpo}{HTML}{4DBBD5}
\definecolor{octopus}{HTML}{669999}
\newcolumntype{L}{>{\hspace{2.2pt}}l<{\hspace{2.2pt}}}
\newcolumntype{Y}{>{\hspace{2.2pt}\centering\arraybackslash}X<{\hspace{2.2pt}}}

\usepackage{microtype}

\usepackage{inconsolata}

\usepackage{graphicx}

\title{Instructions for *ACL Proceedings}

\author{
Yongyao Wang$^{1}$\thanks{Equal contribution.},
Ziqi Miao$^{1}$\footnotemark[1],
Lu Yang$^{2}$,
Haonan Jia$^{3}$
\\
\textbf{
Wenting Yan$^{4}$,
Chen Qian$^{5}$,
Lijun Li$^{1}$\thanks{Corresponding author.}
}
\\
$^1$ Shanghai Artificial Intelligence Laboratory \\
$^2$ The Hong Kong Polytechnic University \\
$^3$ Beihang University \\
$^4$ Zhejiang University \\
$^5$ Renmin University of China \\
\texttt{\{lilijun\}@pjlab.org.cn}
}

\begin{document}

\def\name{TabSieve\xspace}
\def\sftname{\name-SFT-40K\xspace}
\def\rlname{\name-RL-40K\xspace}
\title{TabSieve: Explicit In-Table Evidence Selection for Tabular Prediction}

\maketitle

\begin{figure*}
  \centering
  \includegraphics[width=\linewidth]{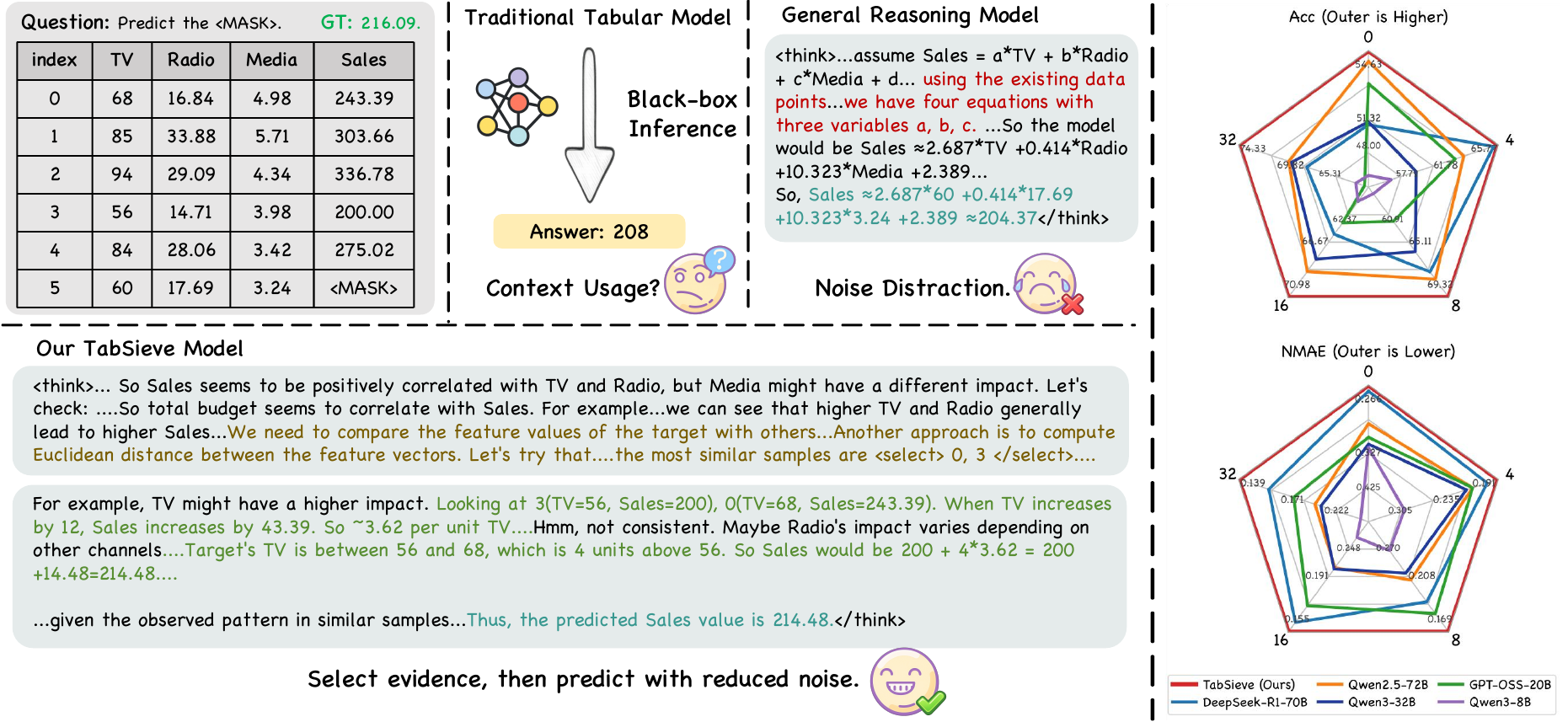}
\vspace{-0.6cm}
\caption{We contrast three prediction paradigms. Traditional models cannot explicitly interpret how context is used. Strong LLMs with in-context prompting can be distracted by \textcolor[HTML]{D62728}{noisy context}. TabSieve first performs \textcolor[HTML]{8C564B}{evidence selection} and then conducts \textcolor[HTML]{2CA02C}{noise-filtered reasoning} to produce the final \textcolor[HTML]{17BECF}{prediction}.}
% \caption{\textbf{Left:} We contrast three prediction paradigms. Traditional models cannot explicitly interpret how context is used. Strong LLMs with in-context prompting can be distracted by \textcolor[HTML]{D62728}{noisy context}. TabSieve first performs \textcolor[HTML]{8C564B}{evidence selection} and then conducts \textcolor[HTML]{2CA02C}{noise-filtered reasoning} to produce the final \textcolor[HTML]{17BECF}{prediction}.
% \textbf{Right:} We compare TabSieve with strong LLM baselines across multiple few-shot settings on both classification and regression tasks.}
  \label{fig:intro}
  \vspace{-0.2cm}
\end{figure*}

\begin{abstract}
  Tabular prediction can benefit from in-table rows as few-shot evidence, yet existing tabular models typically perform instance-wise inference and LLM-based prompting is often brittle. Models do not consistently leverage relevant rows, and noisy context can degrade performance. To address this challenge, we propose TabSieve, a select-then-predict framework that makes evidence usage explicit and auditable. Given a table and a query row, TabSieve first selects a small set of informative rows as evidence and then predicts the missing target conditioned on the selected evidence. To enable this capability, we construct TabSieve-SFT-40K by synthesizing high-quality reasoning trajectories from 331 real tables using a strong teacher model with strict filtering. Furthermore, we introduce TAB-GRPO, a reinforcement learning recipe that jointly optimizes evidence selection and prediction correctness with separate rewards, and stabilizes mixed regression and classification training via dynamic task-advantage balancing. Experiments on a held-out benchmark of 75 classification and 52 regression tables show that TabSieve consistently improves performance across shot budgets, with average gains of 2.92\% on classification and 4.56\% on regression over the second-best baseline. Further analysis indicates that TabSieve concentrates more attention on the selected evidence, which improves robustness to noisy context. 
\end{abstract}

\section{Introduction}
% Tabular data are pervasive in applications such as climate science, healthcare, finance and energy~\cite{cachay2021climart,liu2025interpretable,addo2018credit,colverd2025machine}, and constitute one of the most prevalent formats for structured information. 
Tabular analysis serves as the essential diagnostic foundation for understanding structured data in applications such as climate science, healthcare, finance and energy~\cite{cachay2021climart,liu2025interpretable,addo2018credit,colverd2025machine}, while tabular prediction transforms those historical insights into actionable foresight to drive high-stakes decision-making. Unlike isolated instance prediction, real-world tables typically contain other rows that are statistically correlated with the target row. Therefore, in-context learning (ICL) in tabular prediction is vital to achieve robust generalization and stable performance.
% Such rows naturally serve as in-table evidence for prediction. Accordingly, a model’s ability to leverage such evidence is critical to achieving robust generalization and stable performance.

However, most existing methods still perform instance-wise inference~\cite{tabnet,tabtransformer,fttrans}. 
%The predictor conditions only on the target row, and information from the remaining rows is captured implicitly through parameters learned during training. 
This design becomes brittle when transferring across tables whose schemas, column semantics, or data distributions differ. Although recent foundation models incorporate table context, they remain constrained by limited reasoning capabilities for textual metadata and high sensitivity to context composition.
%Recent in-context learning tabular foundation models~\cite{tabpfn, tabicl} incorporate table context during training and inference, yet two challenges persist. First, they typically offer limited support for reasoning over textual metadata such as column and table descriptions, which constrains ability to incorporate semantic priors and general knowledge. Second, their performance is sensitive to the amount and composition of context, which is particularly pronounced in few-shot scenario.
Conversely, Large Language Models (LLMs) exhibit strong reasoning and in-context learning capabilities~\cite{gpt,deepseek} to cast tabular prediction as language modeling by serializing headers and rows into natural language~\cite{p2t, tabllm, gtl}. Naively concatenating a few-shot set into the prompt does not reliably yield controllable gains in practice. Two risks are particularly salient. (i) The model may not reliably condition its predictions on the provided context during inference, particularly for non-reasoning models~\cite{cheng2025revisiting,sun2025lost}. (ii) Recent studies suggest that many failures of strong LLMs arise from ignoring contextual details or incorrectly applying the context~\cite{clbench}. In realistic deployments, available demonstrations are often noisy or weakly relevant. The lack of a mechanism to select informative rows can cause the model to rely on unhelpful evidence, resulting in degraded accuracy and robustness~\cite{gao2024noise}.
% Without a mechanism to choose informative rows, the model may rely on unhelpful evidence and suffer degraded accuracy and robustness~\cite{zhang2024noise}.

To address these issues, we propose \name, a \emph{select-then-predict} reasoning framework for tabular prediction. As shown in Figure~\ref{fig:intro}, given a table and a query row, \name first analyzes the column semantics to form an initial hypothesis about the relationship between input features and the target. It then explicitly selects a small set of evidential rows that are most informative for the query and finally predicts the missing value. Compared to passively appending demonstrations, \name turns row selection into a constrained intermediate decision, enabling the model to prioritize high-value evidence under a limited context budget and reducing sensitivity to noisy rows.

To equip the model with these capabilities, we build a rigorous data synthesis pipeline that converts $331$ tables from existing collections~\cite{gtl,tpberta} into a large-scale set of training instances with intermediate reasoning traces. Specifically, we construct a teacher-driven workflow that performs structure understanding and feature analysis, explicitly selects effective reference rows, and then completes the final prediction. 
By applying strict filtering and rejection sampling to retain only high-quality trajectories, we construct a cold-start supervised dataset \sftname that provides direct supervision for both evidence selection and prediction.
% We apply strict filtering and rejection sampling to retain only high quality trajectories, yielding a cold-start supervised dataset \sftname that provides direct supervision for both evidence selection and prediction.

Building on this dataset, we further introduce TAB-GRPO for reinforcement learning (RL) by designing separate rewards for evidence selection and prediction to jointly improve selection quality and answer accuracy. In addition, a practical complication in tabular prediction is the coexistence of classification and regression objectives. We observe that in early training, regression tasks often exhibit larger advantage scales, which can dominate the optimization direction and induce early-stage bias. To mitigate this effect, we propose a task-advantage balancing mechanism to adaptively rescale the regression advantages. This rescaling dampens overly strong updates driven by regression at the beginning of training, leading to more stable joint optimization. Our contributions are summarized as follows:
\begin{itemize}
    \item We propose \name, a reasoning-based tabular predictor that explicitly selects in-table evidence and uses it to support prediction. We also construct \sftname to provide high-quality supervision.
    \item We introduce TAB-GRPO, a dynamic task-advantage balancing strategy, to mitigate early-stage optimization imbalance between classification and regression.
    % in tabular reinforcement learning.
    \item We evaluate \name on 75 classification tables and 52 regression tables. \name consistently outperforms baselines, yielding average gains of $2.92\%$ on classification and $4.56\%$ on regression over the second-best method. 
    % Further analyses show that context selection increases attention to the selected evidence.

\end{itemize}

\section{Diagnosing the Need to Optimize Tabular In-Context Learning}
Previous studies have investigated the role of attention in hallucination and grounding failures, suggesting that insufficient attention allocation to evidential context is a potential factor behind ungrounded outputs~
\cite{huang2024opera,huo2024self,liu2025more}. At the same time, several studies indicate that tokens with higher attention weights tend to have a stronger influence on the final decision~\cite{eraser,eilertsen2025aligning}. Motivated by these observations, in this section, we use attention-based analysis to examine how general LLMs utilize tabular context during prediction. Specifically, we center our study on two research questions:

\textbf{RQ1:} Can introducing an explicit evidence selection declaration in the reasoning trace steer the model’s attention toward evidential rows?

\textbf{RQ2:} If the model is guided to emphasize noisy context, can the context become actively misleading and drive the model to incorrect answers?

\begin{figure}[!t]
  \centering
  \includegraphics[width=\linewidth]{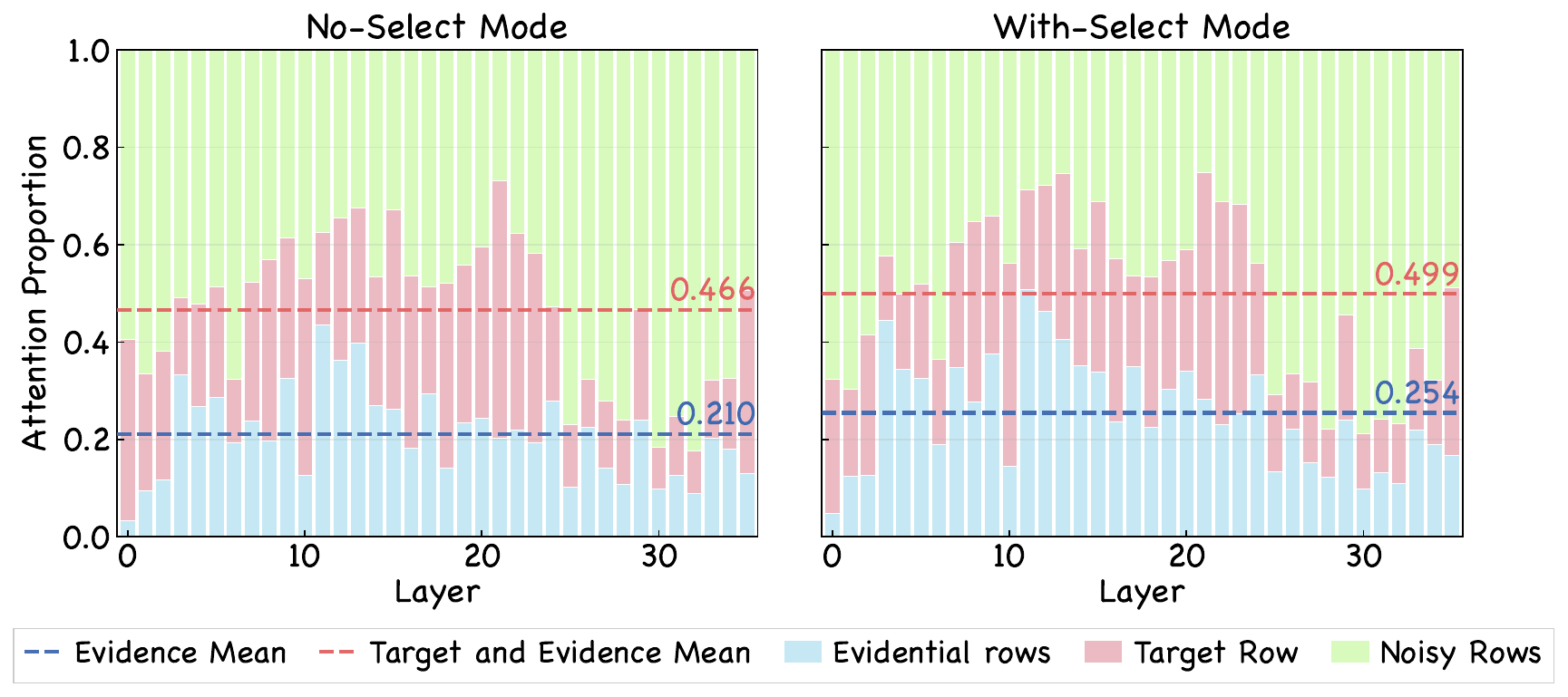}
  \vspace{-0.7cm}
  \caption{Explicit evidence selection concentrates attention on informative rows.}
  \vspace{-0.3cm}
  \label{fig:attn_select}
\end{figure}

\begin{figure}[!t]
  \centering
  \includegraphics[width=\linewidth]{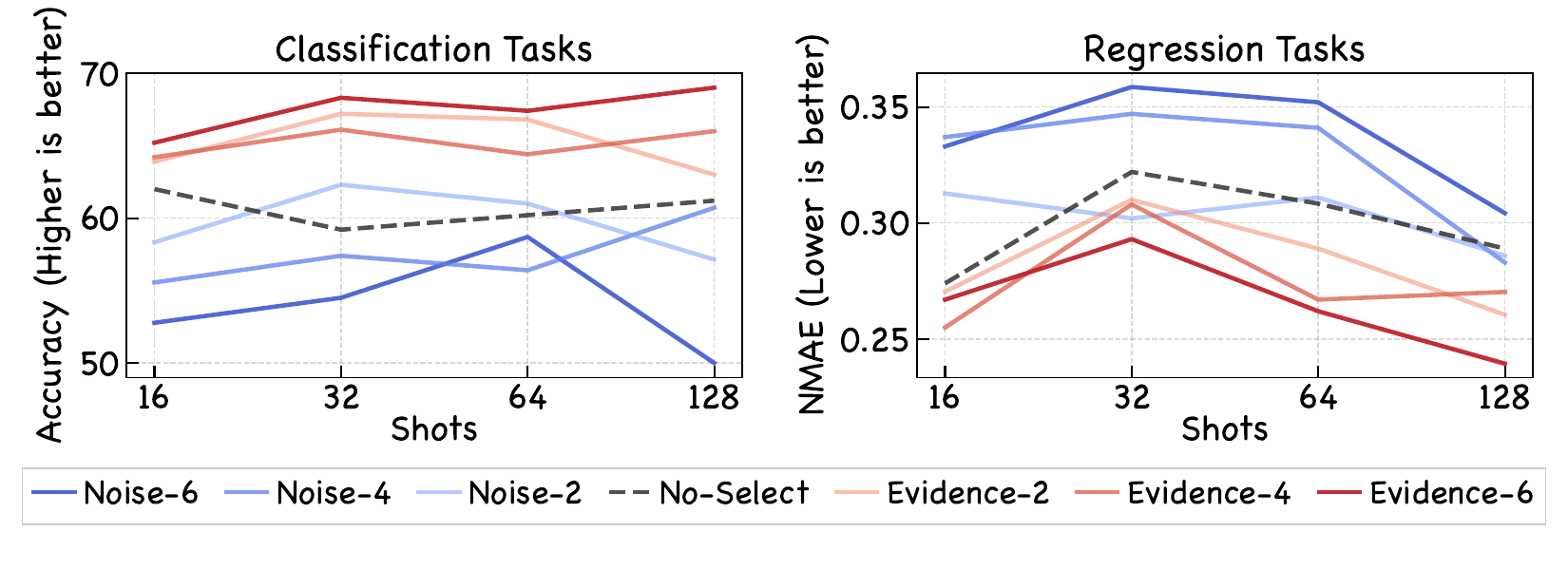}
  \vspace{-0.9cm}
  \caption{Effect of explicit row selection under evidential or noisy in-table context.}
  \vspace{-0.2cm}
  \label{fig:shot_scaling}
\end{figure}

\subsection{Evidence Focus Improves with an Explicit Selection Trace}
Prior work~\cite{thomas2024retrieval} shows that training tabular models with the k-nearest neighbours of a target row can improve accuracy. Inspired by this observation, we categorize the table rows into target row, evidential rows, and noisy rows. The target row is the query instance. Evidential rows are defined as the rows with high embedding similarity to the target row, while the remaining context rows are treated as noise. We then conduct a comparative analysis of the attention allocation of Qwen3-8B~\cite{qwen3} to these token groups under two inference modes. In the With-Select mode, we insert a short \texttt{<select>} block into the reasoning trace, listing the indices of the evidential rows that the model will use for prediction. In the No-Select mode, the model performs standard inference and outputs the answer directly.

\textbf{An explicit \texttt{<select>} declaration shifts attention toward evidential rows and away from noisy context.}
Figure~\ref{fig:attn_select} shows the layer-wise attention proportions over table tokens under the two inference modes. After inserting \texttt{<select>}, the model allocates more attention to evidential rows and less attention to noisy rows. This suggests that explicit selection can serve as an effective intermediate scaffold for guiding evidence-grounded prediction.

\subsection{Noisy Context Can Be Actively Misleading Under In-Context Learning}
\label{sec:diag}
We further examine whether incorrect evidence selection can actively mislead tabular prediction, rather than being merely unhelpful. For each shot setting, we sample 80 prediction tasks. Within each task, we label in-context rows as evidence or noise using the same embedding-similarity criterion. We then compare three inference settings that differ only in the inserted selection trace: (i) No-Select, which performs standard inference without an explicit selection trace; (ii) Evidence-$S$, which inserts a \texttt{<select>} trace listing $S$ evidential rows; and (iii) Noise-$S$, which inserts the same trace but lists $S$ rows sampled from the noisy set.

\textbf{Selecting noisy rows degrades performance, showing that incorrect evidence focus can actively mislead prediction.}
As shown in Figure~\ref{fig:shot_scaling}, Evidence-$S$ generally outperforms No-Select, whereas Noise-$S$ hurts performance, with the gap typically widening as $S$ increases. These results highlight the importance of explicitly optimizing the model's ability to identify valid evidence, as attending to noisy rows can bias inference and reduce prediction quality. At the same time, the gains from Evidence-$S$ further suggest that the embedding-based criterion provides a useful proxy for rows that improve prediction.

\begin{figure*}[!t]
  \centering
  \includegraphics[width=\linewidth]{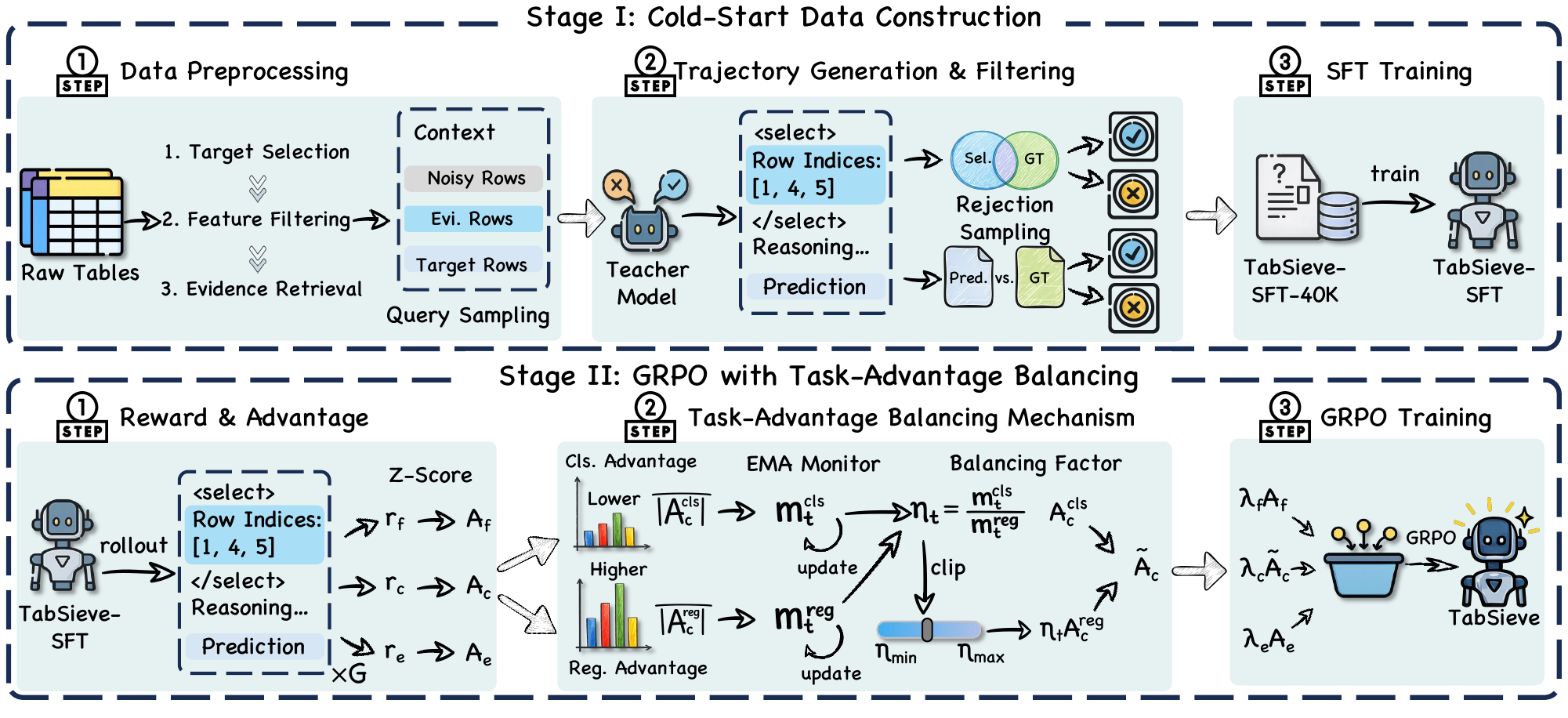}
  \vspace{-0.7cm}
\caption{Training pipeline of TabSieve. We synthesize select-then-predict trajectories from a teacher model to build SFT dataset. Starting from the SFT-initialized model, TAB-GRPO extends GRPO with task-advantage balancing to mitigate optimization imbalance and strengthen context selection for robust in-context learning.}
    \vspace{-0.1cm}
  \label{fig:method}
\end{figure*}

\section{Method}
The above analysis shows that tabular ICL exhibits weak evidence focus by default, while inserting an explicit \texttt{<select>} trace can shift attention from noisy rows toward evidential rows. However, if the model mistakenly treats noisy rows as evidence, prediction performance degrades.
To address these issues, we propose a training strategy to train \textsc{\name}, as illustrated in Figure~\ref{fig:method}. We first conduct cold-start initialization on \sftname, a synthesized dataset of select-then-predict trajectories. We then perform RL with TAB-GRPO to jointly refine evidence selection and improve prediction accuracy. To stabilize early-stage optimization, TAB-GRPO balances task advantages across regression and classification objectives.

\subsection{Preliminary}
\paragraph{Tabular Prediction Tasks.}
We consider a labeled tabular dataset $\mathcal{D}=\{(x_i,y_i)\}_{i=1}^{N}$ with $N$ instances and $d$ feature columns.
Each row $x_i=(x_{i1},\ldots,x_{id})$ lies in the instance space
$\mathcal{X}=\mathcal{X}_1\times\cdots\times\mathcal{X}_d$ and $y_i\in\mathcal{Y}$ is the target.
For $C$-class classification, $\mathcal{Y}=\{0,1,\ldots,C-1\}$; for regression, $\mathcal{Y}\subseteq\mathbb{R}$.
The tabular prediction problem is to learn a predictor $f_{\theta}:\mathcal{X}\rightarrow\mathcal{Y}$
such that, given a query row $x\in\mathcal{X}$, the model outputs $\hat{y}=f_{\theta}(x)$.

% \subsubsection{Tabular Prediction Tasks.}
% We consider a labeled tabular dataset $\mathcal{D}=\{(x_i,y_i)\}_{i=1}^{n}$, where each row
% $x_i=(x_{i1},\ldots,x_{id})$ lies in the instance space
% $\mathcal{X}=\mathcal{X}_1\times\cdots\times\mathcal{X}_d$ and $y_i\in\mathcal{Y}$ is the target.
% For $C$-class classification, $\mathcal{Y}=\{0,1,\ldots,C-1\}$; for regression, $\mathcal{Y}\subseteq\mathbb{R}$.
% The tabular prediction problem is to learn a predictor $f_{\theta}:\mathcal{X}\rightarrow\mathcal{Y}$
% such that, given a query row $x\in\mathcal{X}$, the model outputs $\hat{y}=f_{\theta}(x)$.

\paragraph{In-Context Learning for Tabular Data.}
In the ICL setting, we test on a new tabular task with dataset $\mathcal{D}'$.
Besides a query instance $x^{(q)}\in\mathcal{X}'$, the model is provided with a context set of $K$ labeled
examples $\mathcal{C}=\{(x^{(k)},y^{(k)})\}_{k=1}^{K}\subseteq\mathcal{X}'\times\mathcal{Y}'$,
typically drawn from the same table as the query.
The model predicts by conditioning on the context $\hat{y}^{(q)} = f_{\theta'}\!\big(x^{(q)}, \mathcal{C}\big).$
Equivalently, an ICL predictor implements a mapping
$f_{\theta'}:(\mathcal{X}'\times\mathcal{Y}')^{K}\times\mathcal{X}'\rightarrow\mathcal{Y}'$ and adapts to $\mathcal{D}'$ without updating parameters.

\subsection{Cold-Start Data Construction}\label{sec:sft}
To bootstrap \name with both evidence selection and prediction capability, we construct a cold-start supervised dataset, \sftname, which provides explicit signals for the select-then-predict process. Since existing tabular prediction datasets lack chain-of-thought reasoning traces, we build a reasoning corpus from $331$ raw tables collected by previous methods~\cite{gtl,tpberta}. Specifically, we first preprocess them into task instances, and then apply a teacher-driven workflow to synthesize select-then-predict trajectories.

\paragraph{Data Preprocessing.}
We convert raw tables into task instances through three steps. (i) \textit{Target selection.} We prompt a teacher to evaluate each column as a candidate target based on its name semantics and inferability from the remaining columns with self-consistency. This yields reverse reasoning tasks that map original targets back to the features, improving data diversity~\cite{reversethink}. (ii) \textit{Weak correlation feature filtering.} We first remove identifier columns, then rank the remaining features by mutual information with the target and retain the minimal feature subset whose cumulative mutual information exceeds $90\%$, thereby shortening prompt length and improving sample efficiency. (iii) \textit{Evidence retrieval.} Motivated by the observations in Section~\ref{sec:diag} and findings in prior work~\cite{thomas2024retrieval}, we remove the target column and retrieve the top-32 nearest neighbors of the query row as the evidence pool using the cosine similarity of row embeddings. We then randomly draw $K\in\{0,4,8,16,32\}$ rows from the evidence pool and the full table as the in-context support set, while keeping the evidence ratio within $[0,0.5]$ to prevent the task from becoming trivially easy.

\paragraph{Trajectory Generation and Filtering.}
Given a preprocessed task, a teacher model synthesizes a select-then-predict trajectory through two steps. (i) \textit{Evidence selection.} We ask the teacher to analyze the feature relationships in the table and select $E$ evidential rows among the $K$ candidates. We then prompt the teacher to rewrite the selection trace, removing any statements that indicate prior knowledge of $E$, so that the final trajectory does not leak this information. (ii) \textit{Target prediction.} Starting from the accepted selection trace, the teacher performs up to five attempts at predicting the target value. For classification tasks, we keep only samples whose predictions exactly match the ground truth. For regression tasks, we accept predictions with $\mathrm{MAPE}<25\%$ and further prompt the teacher to refine its reasoning so that the final value matches the ground truth exactly. Finally, we remove trajectories with abnormal length and apply an LLM-as-judge to filter out traces with logical errors, implausible reasoning, and potential answer leakage. Distribution of the resulting \sftname is detailed in Appendix~\ref{sec:distru}.

\subsection{GRPO with Task-Advantage Balancing}
We further strengthen \name using Group Relative Policy Optimization (GRPO)~\cite{deepseekmath} to jointly improve evidence selection and target prediction. In addition, we introduce a task-advantage balancing mechanism to alleviate the early-training imbalance where regression tends to dominate gradient updates, thereby stabilizing joint learning over classification and regression.

\paragraph{Reward Design.}
Each training instance is a quadruple $(x,y,E^\star,\tau)$, where $x$ is the prompt, $y$ is the ground-truth target, $E^\star$ is the pseudo-evidence set, and $\tau\in\{\mathrm{cls},\mathrm{reg}\}$ indicates the task type. Given a sampled output $o$, we define three rewards. The \emph{format reward} $r_f\in\{0,1\}$ verifies whether $o$ contains a valid \texttt{<select>} block and a \texttt{\textbackslash boxed\{$\cdot$\}} answer. The \emph{evidence reward} $r_e$ is the $F_1$ score between the selected row indices $\hat{E}$ parsed from \texttt{<select>} and the pseudo set $E^\star$. The \emph{correctness reward} $r_c$ evaluates prediction quality in a task-dependent manner. For classification, we use exact match between the prediction and the ground truth, i.e., $r_c=\mathbf{1}[\hat{y}=y]$. For regression, we apply an exponential shaping on the mean absolute percentage error, i.e., $r_c=\exp(-\gamma\cdot\mathrm{MAPE}(\hat{y},y))$ with $\mathrm{MAPE}(\hat{y},y)=|\hat{y}-y|/|y|$.

\paragraph{Advantage Computation.}
For each prompt $x$, we sample a group of $G$ outputs $\{o_i\}_{i=1}^{G}$ from the behavior policy $\pi_{\theta^-}$ and, for each reward component $k\in\{f,e,c\}$, compute a group-relative advantage by whitening rewards within the group:
\begin{equation}
A^{(k)}_i=\frac{r^{(k)}_i-\mathrm{mean}\big(\{r^{(k)}_j\}_{j=1}^{G}\big)}{\mathrm{std}\big(\{r^{(k)}_j\}_{j=1}^{G}\big)}.
\label{eq:group_whiten}
\end{equation}

\paragraph{Task-Advantage Balancing Mechanism.}
\textit{Motivation.} In mixed-task training, regression samples exhibit a larger correctness-advantage magnitude $|A^{(c)}|$ than classification samples in the early stage. Since GRPO weights policy updates by advantages, this imbalance lets regression dominate the update direction and may prematurely bias the model toward a regression local optimum before robust cross-task behaviors emerge.

\textit{Balancing factor.} To mitigate this, we track the task-wise advantage scale via exponential moving averages (EMA) of $|A^{(c)}|$ with decay $\beta\in(0,1)$:
\begin{equation}
m_t^{\tau}=\beta\,m_{t-1}^{\tau}+(1-\beta)\,\mathbb{E}\!\left[|A^{(c)}|\,\big|\,\tau\right].
\label{eq:ema_adv}
\end{equation}
We then construct a bounded balancing factor that only rescales regression advantages:
\begin{equation}
\eta(\tau)=
\begin{cases}
1, & \tau=\mathrm{cls},\\
\mathrm{clip}\!\big(\frac{m_t^{\mathrm{cls}}}{m_t^{\mathrm{reg}}},\,\eta_{\min},\,\eta_{\max}\big), & \tau=\mathrm{reg},
\end{cases}
\label{eq:task_scaler}
\end{equation}
and apply it to obtain the balanced correctness advantage $\widetilde{A}^{(c)}_i=\eta(\tau)\,A^{(c)}_i$, where $0<\eta_{\min}\le\eta_{\max}\le 1$. This dampens overly strong regression updates and narrows the optimization gap between the two task types.

\paragraph{GRPO Objective.}
We aggregate the three component advantages into a single advantage $A_i=\lambda_f A^{(f)}_i+\lambda_e A^{(e)}_i+\lambda_c \widetilde{A}^{(c)}_i$, where $\lambda_f,\lambda_e,\lambda_c$ control their relative importance. Following GRPO, we optimize:
\begin{equation}
\begin{aligned}
\max_{\theta}\;&\mathbb{E}\Bigg[\frac{1}{G}\sum_{i=1}^{G}\min\Big(\frac{\pi_{\theta}(o_i\mid x)}{\pi_{\theta^{-}}(o_i\mid x)}A_i, \\
&\mathrm{clip}(\frac{\pi_{\theta}(o_i\mid x)}{\pi_{\theta^{-}}(o_i\mid x)},1-\epsilon_{low},1+\epsilon_{high})A_i\Big)\Bigg],
\end{aligned}
\label{eq:tab_grpo_obj}
\end{equation}
where $\epsilon_{low}$ and $\epsilon_{high}$ are the clipping thresholds.

\begin{table*}[!t]
\centering
\begingroup
\scriptsize
\setlength{\tabcolsep}{1.5pt}
\renewcommand{\arraystretch}{1}
\setlength{\aboverulesep}{0.2ex}
\setlength{\belowrulesep}{0.2ex}
\begin{tabularx}{\textwidth}{ >{\raggedright\arraybackslash}p{0.27\textwidth} *{5}{Y} c c *{5}{Y} c c }
\toprule
 & \multicolumn{7}{c}{\textbf{Classification (Accuracy $\uparrow$)}} & \multicolumn{7}{c}{\textbf{Regression (NMAE $\downarrow$)}} \\
\cmidrule(lr){2-8} \cmidrule(lr){9-15}
\textbf{Model} & \textbf{0} & \textbf{4} & \textbf{8} & \textbf{16} & \textbf{32} & \textbf{Rank$_z$} & \textbf{Rank$_i$} & \textbf{0} & \textbf{4} & \textbf{8} & \textbf{16} & \textbf{32} & \textbf{Rank$_z$} & \textbf{Rank$_i$} \\
\midrule

\rowcolor{gspo!10}\multicolumn{15}{c}{\textbf{Traditional Tabular Models}} \\
\midrule
XGBoost~\cite{xgboost}           & -- & 49.93 & 51.79 & 58.38 & 62.98 & --   & 10.63 & --    & 0.237 & 0.200 & 0.184 & 0.160 & --    & 8.23  \\
CatBoost~\cite{catboost}         & -- & 51.25 & 56.42 & 65.88 & 72.64 & --   & 8.95  & --    & 0.235 & 0.192 & 0.179 & 0.159 & --    & 7.56  \\
FT-Transformer~\cite{fttrans}    & -- & 54.31 & 59.74 & 68.60 & 71.45 & --   & 8.37  & --    & 0.363 & 0.363 & 0.357 & 0.354 & --    & 11.56 \\
TabPFN~\cite{tabpfn}             & -- & 52.68 & 60.70 & \underline{69.15} & \underline{73.67} & -- & 8.09 & -- & 0.228 & 0.186 & 0.165 & \underline{0.144} & -- & 6.70 \\
\midrule

\rowcolor{gspo!10}\multicolumn{15}{c}{\textbf{General Large Language Models}} \\
\midrule
Qwen2.5-72B-Instruct~\cite{qwen2.5}     & \underline{54.07} & 63.67 & \underline{67.95} & 68.98 & 70.87 & \underline{6.00} & \underline{6.12} & 0.297 & 0.206 & 0.205 & 0.198 & 0.178 & 5.19 & 7.15 \\
DeepSeek-R1-70B~\cite{deepseek}         & 51.04 & \underline{65.52} & 67.39 & 66.06 & 69.62 & 6.37 & 6.76 & \underline{0.270} & \underline{0.197} & 0.188 & \underline{0.160} & 0.152 & \underline{4.81} & \underline{6.40} \\
QwQ-32B~\cite{qwq32b}                   & 52.07 & 64.11 & 62.46 & 65.62 & 67.07 & 6.59 & 8.28 & 0.293 & 0.220 & 0.196 & 0.164 & 0.158 & 4.90 & 6.61 \\
Qwen3-32B~\cite{qwen3}                  & 51.18 & 60.74 & 65.84 & 68.01 & 70.67 & 6.16 & 6.17 & 0.317 & 0.210 & 0.211 & 0.197 & 0.182 & 6.52 & 8.33 \\
GPT-OSS-20B~\cite{gptoss}               & 53.05 & 63.14 & 63.53 & 65.17 & 65.59 & 6.39 & 8.27 & 0.310 & 0.206 & \underline{0.180} & 0.170 & 0.165 & 5.33 & 7.21 \\
GLM-4-32B-0414~\cite{glm4}              & 54.06 & 63.73 & 66.45 & 68.21 & 68.99 & 6.27 & 6.93 & 0.303 & 0.214 & 0.182 & 0.182 & 0.182 & 6.71 & 7.76 \\
Qwen3-8B~\cite{qwen3}                   & 48.18 & 58.77 & 60.91 & 63.01 & 65.67 & 7.44 & 8.56 & 0.325 & 0.268 & 0.239 & 0.235 & 0.222 & 7.14 & 9.10 \\
Llama-3.1-8B~\cite{llama}      & 38.32 & 46.77 & 48.88 & 48.02 & 47.50 & 10.08 & 13.51 & 0.567 & 0.351 & 0.338 & 0.331 & 0.319 & 12.14 & 14.07 \\
Mistral-7B-Instruct-v0.3~\cite{mistral} & 45.94 & 57.18 & 55.78 & 54.37 & 53.82 & 7.37 & 10.51 & 0.668 & 0.416 & 0.405 & 0.370 & 0.365 & 12.67 & 14.57 \\
\midrule

\rowcolor{gspo!10}\multicolumn{15}{c}{\textbf{Tabular Large Language Models}} \\
\midrule
Tabula-8B~\cite{tabula}                 & 39.92 & 41.77 & 41.62 & 44.13 & 42.70 & 9.84 & 14.45 & 0.648 & 0.420 & 0.395 & 0.382 & 0.445 & 13.00 & 15.63 \\
TableLLM-13B~\cite{zhang2025tablellm}  & 30.09 & 40.20 & 41.60 & 39.74 & 33.31 & 10.44 & 13.02 & 0.709 & 0.476 & 0.581 & 0.571 & 0.541 & 12.43 & 15.94 \\
TableGP2-7B~\cite{tabgpt}              & 36.29 & 41.84 & 43.62 & 45.99 & 47.93 & 10.76 & 14.03 & 0.511 & 0.504 & 0.473 & 0.477 & 0.508 & 10.95 & 16.68 \\
P2T (Qwen3-8B)~\cite{p2t}                               & 48.72 & 59.26 & 61.87 & 63.46 & 66.34 & 7.80 & 8.36 & 0.315 & 0.259 & 0.231 & 0.225 & 0.215 & 6.43 & 9.25 \\
GTL-13B~\cite{gtl}                            & 48.62 & 60.35 & 62.53 & 64.27 & 67.15 & 7.33 & 7.92 & 0.304 & 0.250 & 0.226 & 0.215 & 0.203 & 6.29 & 8.89 \\
\midrule

\rowcolor{mygreen!10}
\name (Ours) & \textbf{54.63} & \textbf{65.77} & \textbf{69.32} & \textbf{70.98} & \textbf{74.33} & \textbf{5.76} & \textbf{5.46} & \textbf{0.266} & \textbf{0.192} & \textbf{0.169} & \textbf{0.158} & \textbf{0.139} & \textbf{3.95} & \textbf{5.95} \\
\rowcolor{mygreen!10}
\textit{$\Delta$ over base} & +6.45 & +7.00 & +8.42 & +7.97 & +8.65 & -1.68 & -3.10 & -0.059 & -0.076 & -0.071 & -0.078 & -0.082 & -3.19 & -3.15 \\
\bottomrule
\end{tabularx}
\endgroup
\vspace{-0.1cm}
\caption{Classification (Accuracy $\uparrow$) and Regression (NMAE $\downarrow$) results averaged over all datasets under zero-shot and few-shot settings. Rank$_z$ denotes the average rank in the zero-shot setting; Rank$_i$ is averaged across all datasets and all few-shot settings. $\Delta$ denotes the performance gain relative to Qwen3-8B.}
\label{tab:main_rst}
\vspace{-0.3cm}
\end{table*}

\section{Experiments}
\subsection{Experimental Setup}

\paragraph{Implementation Details.}
To construct the SFT dataset, we process 331 tables collected from previous works~\cite{gtl,tpberta}.
We use Qwen3-Embedding-8B~\cite{qwen3embedding} to encode each table row and Qwen3-Next-80B-Thinking~\cite{qwen3} as the teacher model to synthesize trajectories, and adopt Qwen3-235B-A22B as the judge model for rejection sampling.
In the SFT stage, we initialize from Qwen3-8B~\cite{qwen3} and fine-tune it with LLaMA-Factory~\cite{llamafactory}. For RL stage, our implementation of TAB-GRPO is based on the VeRL~\cite{verl} framework\footnote{\url{https://github.com/Kazeya27/TabSieve}}.
Hyperparameter settings are reported in Appendix~\ref{sec:hyper}.

\paragraph{Evaluation.}
We evaluate on the benchmark released by GTL and TP-BERTa~\cite{gtl,tpberta}, which contains 75 classification tables and 52 regression tables.
To prevent data leakage, we manually filtered out any training tables that appear in the evaluation benchmark.
Following common tabular in-context learning protocols, we report results under multiple shot budgets~\cite{tabr1,tabllm,p2t}.
For each shot setting, we uniformly sample $5$K query rows to form the evaluation set and construct the in-context support set by sampling rows from the same table with three different random seeds.
We use accuracy (Acc) for classification and normalized mean absolute error (NMAE) for regression.
To reduce evaluation variance in few-shot regression and limit the impact of extreme outliers, we clip the NMAE for each sample at $1.0$ before aggregation~\cite{gtl}.

\paragraph{Baselines.}
Our baselines fall into three categories. (i) \emph{Traditional Tabular Models.} Except for TabPFN, all predictors are fine-tuned on the labeled data for each dataset, so we report their performance starting from the few-shot regime rather than the zero-shot setting. (ii) \emph{General Large Language Models.} We evaluate a diverse set of open-source models spanning different scales, and test them under a unified prompting protocol. (iii) \emph{Tabular Large Language Models.} We follow the prompting templates and input formatting used in the official implementations.

\subsection{Main Results}
Table~\ref{tab:main_rst} presents a comprehensive comparison of \name against all baselines on both classification and regression tasks.  Compared with the second-best baseline, \name yields average gains of 2.92\% on classification and 4.56\% on regression.

\paragraph{Comparison to traditional tabular models.}
Traditional tabular models require task-specific fitting on each table and therefore are not applicable in the zero-shot setting. In contrast, \name performs cross-table reasoning without per-table fine-tuning by analyzing table schemas and leveraging semantic reasoning to infer relationships between features and targets. This enables \name to generalize effectively across heterogeneous tables while achieving the best overall performance.

\paragraph{Comparison to general LLMs.}
\name consistently outperforms all evaluated general LLMs, including models with up to 70B parameters. Compared with the base Qwen3-8B, \name delivers substantial gains on both tasks, and these gains generally become larger as the shot budget increases. This trend suggests that \name can exploit in-context examples more effectively. Its select-then-predict procedure prioritizes evidential rows and suppresses noisy context, leading to more effective context utilization.

\paragraph{Comparison to tabular LLMs.}
Prior tabular LLMs typically lack strong reasoning ability and do not provide an explicit chain-of-thought, which limits their generalization under heterogeneous table schemas and task distributions. In contrast, \name explicitly learns to reason over in-table evidence and to filter contextual rows through a dedicated selection step. This design is effective in in-context learning scenarios where demonstrations may be noisy, resulting in more robust and accurate predictions.

\subsection{Ablation Study}
% In this section, we conduct ablations on the training stages, key RL components, and the evidence selection capability.
In this section, we conduct ablations on the training stages, key RL components, and the evidence selection.
% 写的不清楚：
We report the mean results averaged over zero-shot and all few-shot configurations.

\begin{figure}[!t]
  \centering
  \includegraphics[width=\linewidth]{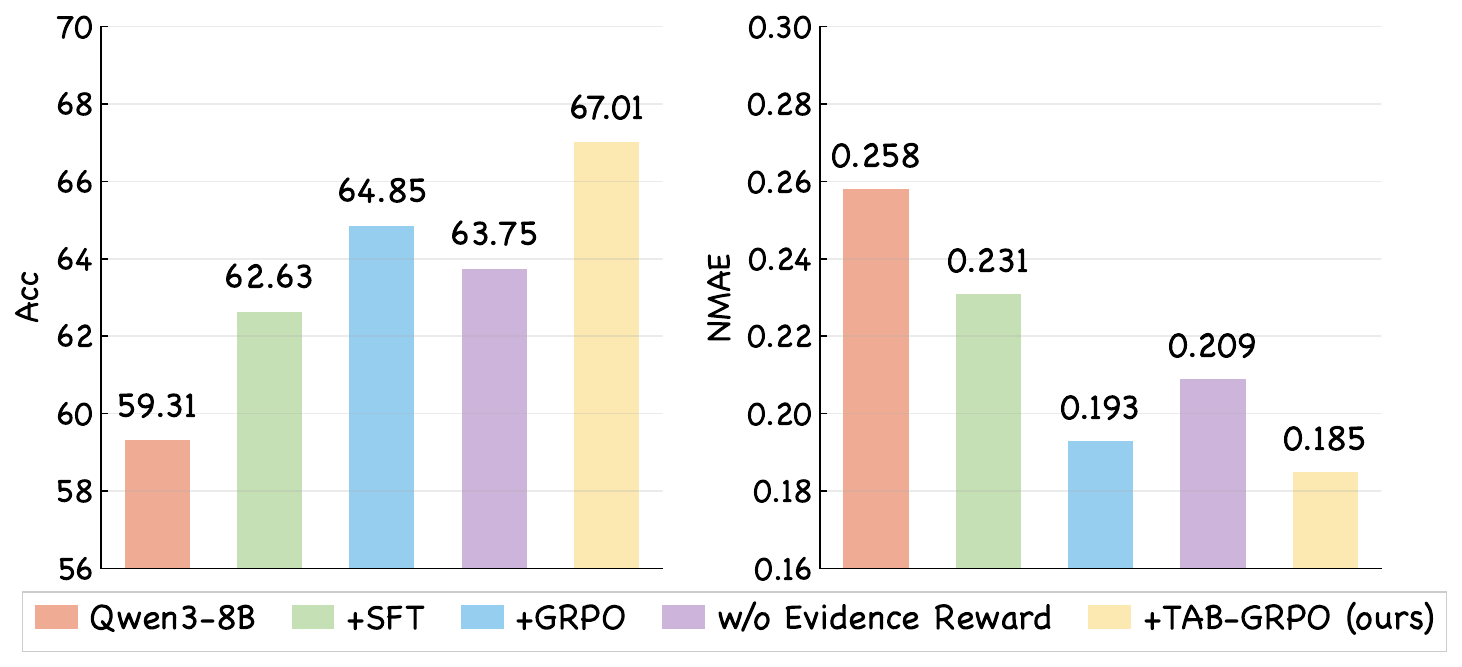}
  \vspace{-0.6cm}
  \caption{Ablation on training stages and RL components.}
  \vspace{-0.2cm}
  \label{fig:ablation1}
\end{figure}

\begin{figure}[!t]
  \centering
  \includegraphics[width=\linewidth]{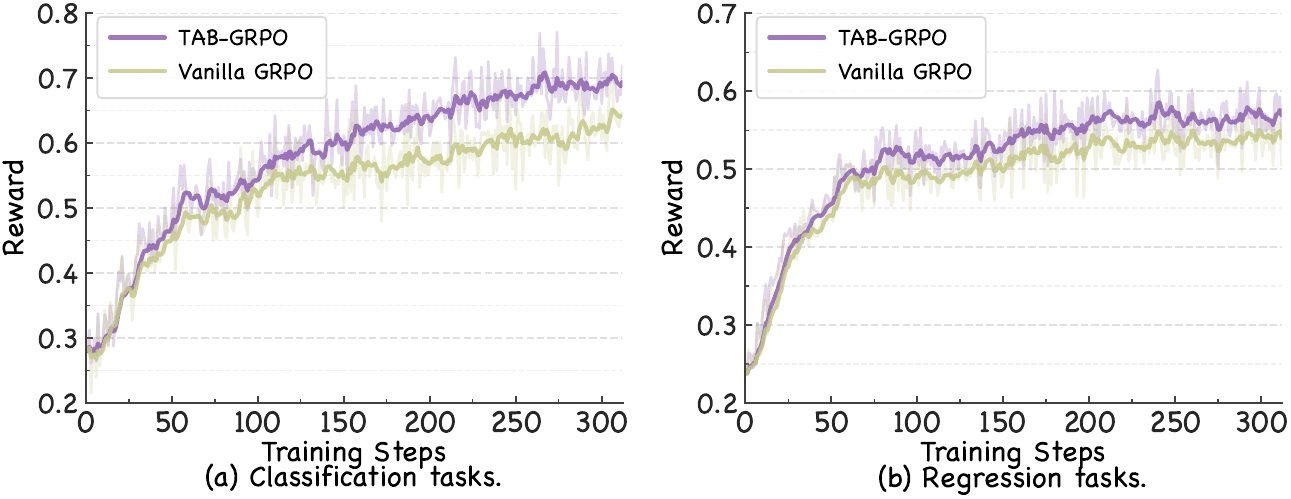}
  \vspace{-0.6cm}
  \caption{Training reward curves of TAB-GRPO and Vanilla GRPO.}
  \vspace{-0.2cm}
  \label{fig:curve}
\end{figure}

\begin{figure}[!t]
  \centering
  \includegraphics[width=\linewidth]{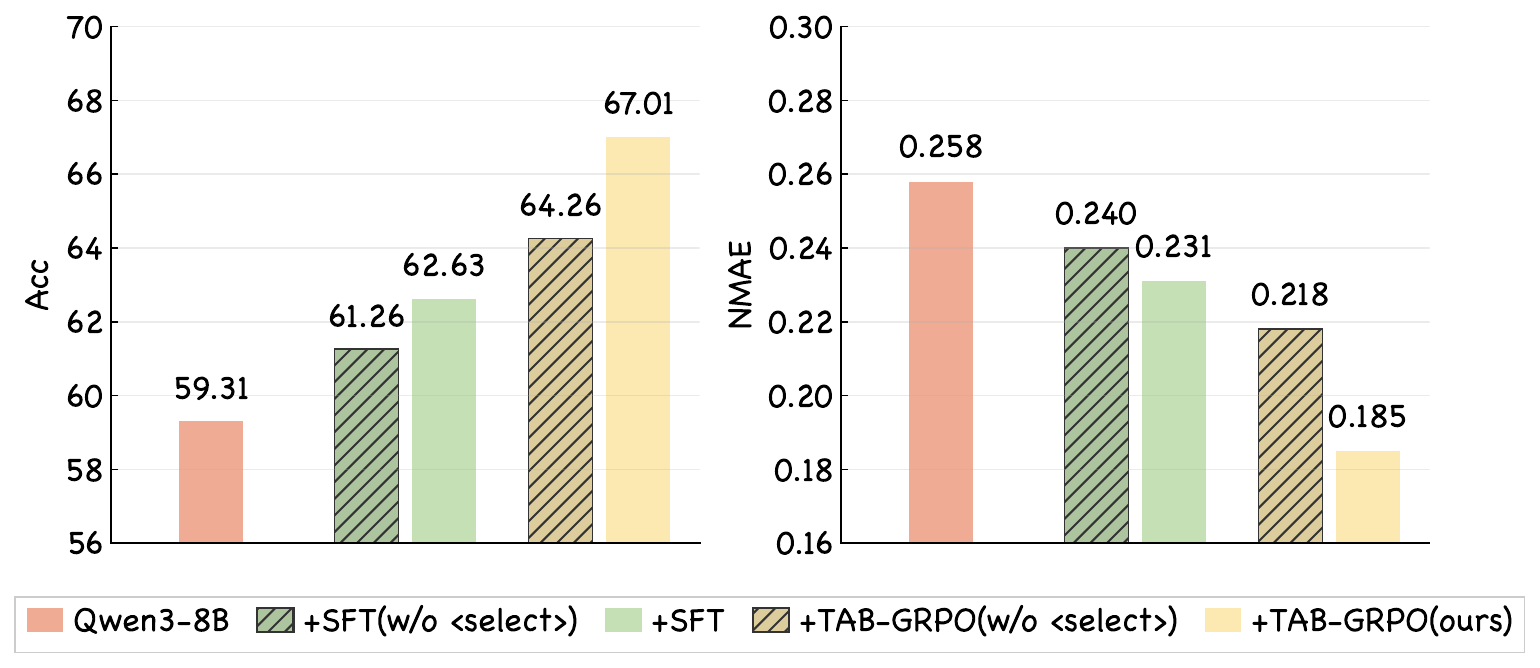}
  \vspace{-0.6cm}
  \caption{Ablation on evidence selection.}
  \vspace{-0.2cm}
  \label{fig:ablation2}
\end{figure}

\paragraph{Training stages.}
Figure~\ref{fig:ablation1} summarizes the incremental gains from SFT and RL stage.
Starting from the backbone \textsc{Qwen3-8B}, SFT on \sftname (\texttt{+SFT}) yields a clear improvement, showing that the synthesized reasoning trajectories provide an effective initialization. Standard GRPO (\texttt{+GRPO}) further improves both tasks, indicating that our reward design provides a reliable optimization signal that effectively refines the selection policy and reinforces prediction quality beyond pure imitation.

\paragraph{RL components.}
Figure~\ref{fig:ablation1} also reports an ablation over key RL components.
Removing the evidence reward (\texttt{w/o Evidence Reward}) leads to a clear degradation, indicating that optimizing evidence selection improves the model's in-context learning ability. In addition, replacing vanilla GRPO with our TAB-GRPO yields a large improvement in classification. We further report the training curves in Figure~\ref{fig:curve}, which show that TAB-GRPO achieves higher rewards during training. This supports our task-advantage balancing design, which reduces early-stage imbalance and stabilizes joint training.
% \paragraph{RL components.}
% Figure~\ref{fig:ablation1} also reports an ablation over key RL components.
% Removing the evidence reward (\texttt{w/o Evidence Reward}) leads to a clear degradation. This result further indicates that explicitly optimizing evidence selection strengthens the model's context-filtering capability, which is essential for improving in-context learning performance beyond answer-level optimization. In addition, replacing vanilla GRPO with our TAB-GRPO yields a large improvement in classification.
% This gain supports our task-advantage balancing design, which effectively alleviates the optimization imbalance in the early-stage training and leads to a more stable joint optimization over classification and regression tasks. 

\paragraph{Evidence selection.}
We construct \texttt{SFT(w/o \texttt{<select>})} by removing the \texttt{<select>} step from the synthesized trajectories and fine-tuning the backbone on the modified dataset. We then continue RL training from \texttt{SFT(w/o \texttt{<select>})} while disabling the evidence reward, yielding \texttt{TAB-GRPO(w/o \texttt{<select>})}. Figure~\ref{fig:ablation2} shows that both variants perform worse than \name at the corresponding training stages. These results indicate that explicit evidence selection is a key capability for unlocking the full benefits of the select-then-predict framework.

% \paragraph{Evidence selection.}
% Figure~\ref{fig:ablation2} quantifies the contribution of explicit evidence selection by removing the selection process from the SFT dataset.
% Specifically, we construct \texttt{SFT(w/o \texttt{<select>})} by deleting the \texttt{<select>} step in the synthesized trajectories and fine-tuning the backbone on this modified dataset.
% The resulting model performs worse than the full SFT model, indicating that the gain is not solely due to supervising the prediction procedure, but also comes from evidence selection.
% This further highlights the importance of evidence selection for in-context learning.
% We then continue RL training on top of \texttt{SFT(w/o <select>)} while disabling the evidence reward, yielding \texttt{TAB-GRPO(w/o \texttt{<select>})}.
% Although RL still improves the model over its SFT initialization, this variant remains consistently weaker than \name.
% These results suggest that explicit evidence selection is a key capability to unlock the full benefits of the select-then-predict framework.

\begin{figure}[!t]
  \centering
  \includegraphics[width=\linewidth]{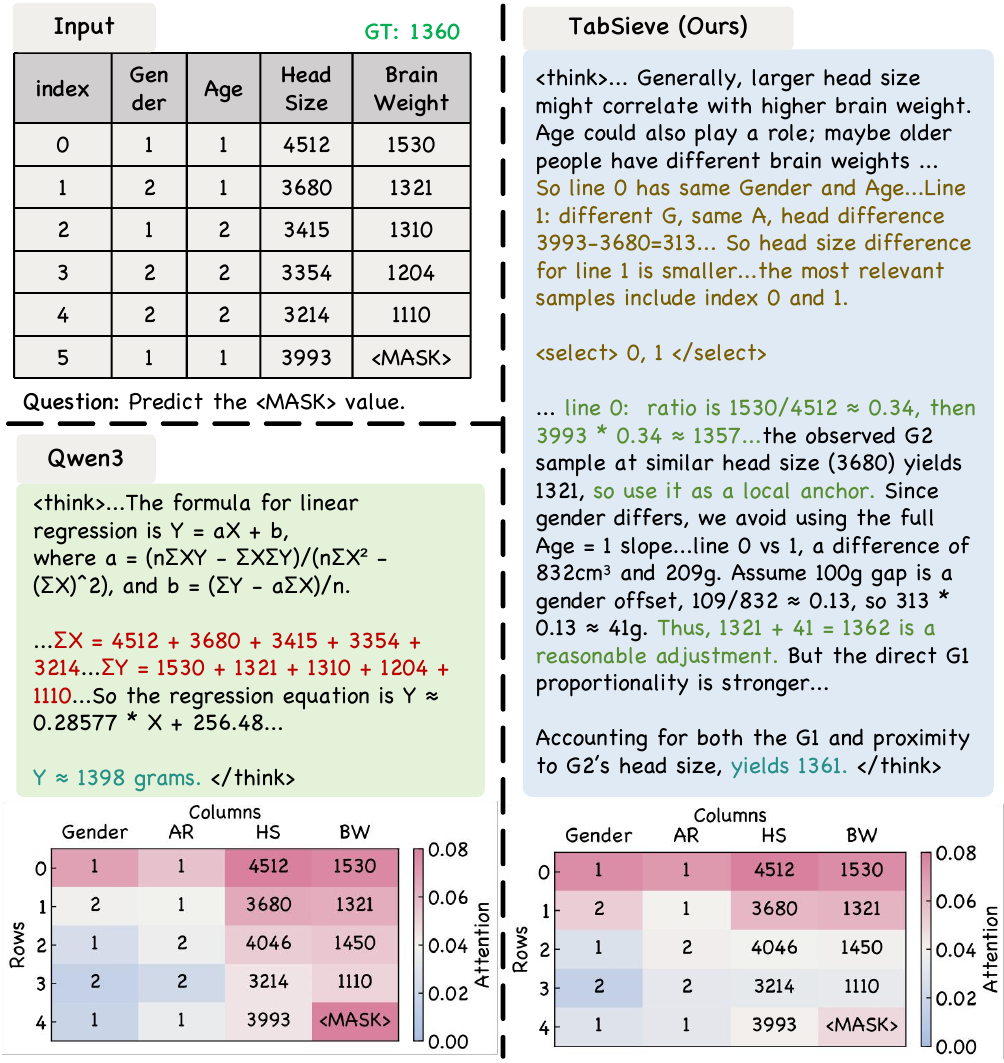}
  \vspace{-0.6cm}
  \caption{Case study of evidence selection.}
  \vspace{-0.3cm}
  \label{fig:case}
\end{figure}

% \subsection{Benefits of Evidence Selection Analysis}

% We provide a case study to highlight how evidence selection improves robustness to noisy in-table context. Figure~\ref{fig:case} compares the Qwen3-8B with TabSieve on a regression instance. Qwen3 performs prediction by conditioning on all contextual rows and inferring a global relationship from the entire support set. This behavior is sensitive to weakly relevant or heterogeneous rows, where spurious correlations in a small context can bias the inferred mapping and lead to inaccurate estimates. In contrast, TabSieve adopts a select-then-predict procedure that first identifies a small subset of informative rows and then performs prediction conditioned on the selected evidence. By filtering out less relevant rows before prediction, TabSieve reduces the influence of noise and encourages locally grounded estimation, resulting in a better prediction.

% We further visualize the aggregated attention over table tokens in the same example. Qwen3 exhibits broadly distributed attention across the context, with high attention on the \textit{TV} feature for nearly every row. In contrast, TabSieve shows a markedly more concentrated attention pattern, allocating much larger attention mass to the selected evidence rows. This alignment between the explicit selection trace and the attention distribution suggests that evidence selection not only improves accuracy, but also makes context usage more interpretable and auditable. Additional case studies are provided in Appendix~\ref{sec:morecase

\subsection{Benefits of Evidence Selection Analysis}
We present a case study in Figure~\ref{fig:case} to illustrate the benefits of explicit evidence selection. Qwen3-8B tends to fit over all contextual rows, which in few-shot settings with sparse context is more easily affected by noisy or weakly relevant rows. In contrast, \name follows a select-then-predict procedure that first identifies informative rows, reduces the influence of noise, and then predicts based on the selected evidence, yielding a better prediction.

We further follow prior work~\cite{cappo} and aggregate attention weights from the final 12 transformer layers into a cell-level heatmap. As shown in Figure~\ref{fig:case}, Qwen3-8B distributes attention broadly across the context, assigning substantial attention to nearly every row and even to the masked target cell, although it contains no task-relevant information. In contrast, \name concentrates attention much more on the selected rows. This suggests that explicit evidence selection not only improves prediction accuracy, but also makes context usage more interpretable and auditable. Additional case studies are provided in Appendix~\ref{sec:more_case}.

\section{Related Works}
\paragraph{Deep Learning Models for Tabular Data.}
Tabular prediction has long been a central problem in machine learning. Gradient-boosted decision trees have historically been strong performers, and prior studies show that tree ensembles often remain highly competitive and outperform many deep models~\cite{shwartz2022tabular,grinsztajn2022tree,stunt}. To narrow this gap, researchers have proposed attention-based neural architectures to better capture feature interactions in tabular data~\cite{tabnet,fttrans,saint}. LoCalPFN~\cite{thomas2024retrieval} shows that training with the k-nearest neighbours of the query row can further improve performance. Beyond single table learning, cross-table generalization methods address schema heterogeneity and knowledge transfer~\cite{transtab,xtab,tpberta}. Despite these advances, most deep tabular models follow an instance-wise inference paradigm and do not exploit within-table evidence during inference. In-context learning models further highlight the potential of context-aware inference, where predictions are produced by conditioning on a set of labeled rows within the table at test time~\cite{tabpfn,tabicl}.

\paragraph{Large Language Models for Tabular Prediction.}
Large language models have recently gained traction for tabular prediction due to their in-context learning and reasoning abilities. Early works serialize tables into text and rely on prompting to elicit predictions~\cite{tabllm,p2t}. Later studies emphasize improving data quality and training signals to better adapt LLMs to heterogeneous tables~\cite{meditab,gtl,tabula}. TabR1 further explores optimization beyond supervised learning by introducing reinforcement learning~\cite{tabr1}. Despite these developments, context usage in existing LLM-based approaches is still largely implicit and heavily dependent on prompt designs, which can be brittle when the context budget is limited and demonstrations are noisy. In contrast, \name makes evidence selection an explicit intermediate action and jointly optimizes it with prediction, enabling more reliable suppression of noisy contextual rows and more consistent use of few-shot evidence in in-context learning.

\section{Conclusion}
In this paper, we propose \name, a \emph{select-then-predict} framework for tabular prediction that makes in-table evidence usage explicit. Rather than passively concatenating demonstrations, \name first selects a small set of informative rows as evidence and then performs prediction conditioned on the selected rows, thereby focusing the context budget on salient evidence and reducing sensitivity to noisy rows. To train this capability, we construct \name-SFT-40K through data synthesis and strict filtering, and further propose TAB-GRPO to stabilize training on mixed regression and classification tasks. Extensive experiments across $127$ tables show that \name consistently improves over strong baselines under different shot budgets, while ablation studies and case studies further validate the effectiveness of its key design choices.

\section*{Limitations}
First, we focus exclusively on text-only tabular prediction and do not investigate whether \name can generalize to multimodal tables that involve images or other non-textual fields. Since many real-world tabular scenarios are inherently multimodal, extending the framework to such settings would be an important direction for future work. Second, due to computational constraints, our current study is limited to the 8B scale. It remains to be explored whether the benefits of \name can be further amplified with larger backbones and broader training budgets. Third, on classification tasks \name only outputs a single predicted class label rather than a probability distribution over the candidate labels, which limits its use in settings that require calibrated class posteriors. We leave extending the output format to per-label probabilities for future work. Finally, our current design mainly optimizes prediction performance. A more general and capable tabular model should ideally also develop stronger table understanding abilities, such as better schema comprehension and structural reasoning. Improving these capabilities could not only broaden the applicability of \name, but also potentially bring further gains in prediction performance.

\section*{Ethical Considerations}
This work aims to improve the prediction performance of tabular LLMs in few-shot settings. All experiments are conducted on publicly available tabular datasets and benchmarks. As in prior work, these data sources and tabular prediction methods may contain social biases, sampling biases, or other imperfections that can affect model behavior and evaluation outcomes. Beyond the risks already associated with tabular prediction methods on existing public datasets, we do not identify additional ethical risks introduced specifically by our method. We encourage continued attention to data quality, fairness, transparent evaluation, and responsible reporting of model capabilities and limitations.

% Bibliography entries for the entire Anthology, followed by custom entries
%\bibliography{anthology,custom}
% Custom bibliography entries only
\bibliography{custom}
\clearpage
\appendix

\twocolumn[{%
\begin{@twocolumnfalse}
\begin{center}
\begin{minipage}{0.98\textwidth}
    \centering
    \includegraphics[width=\linewidth]{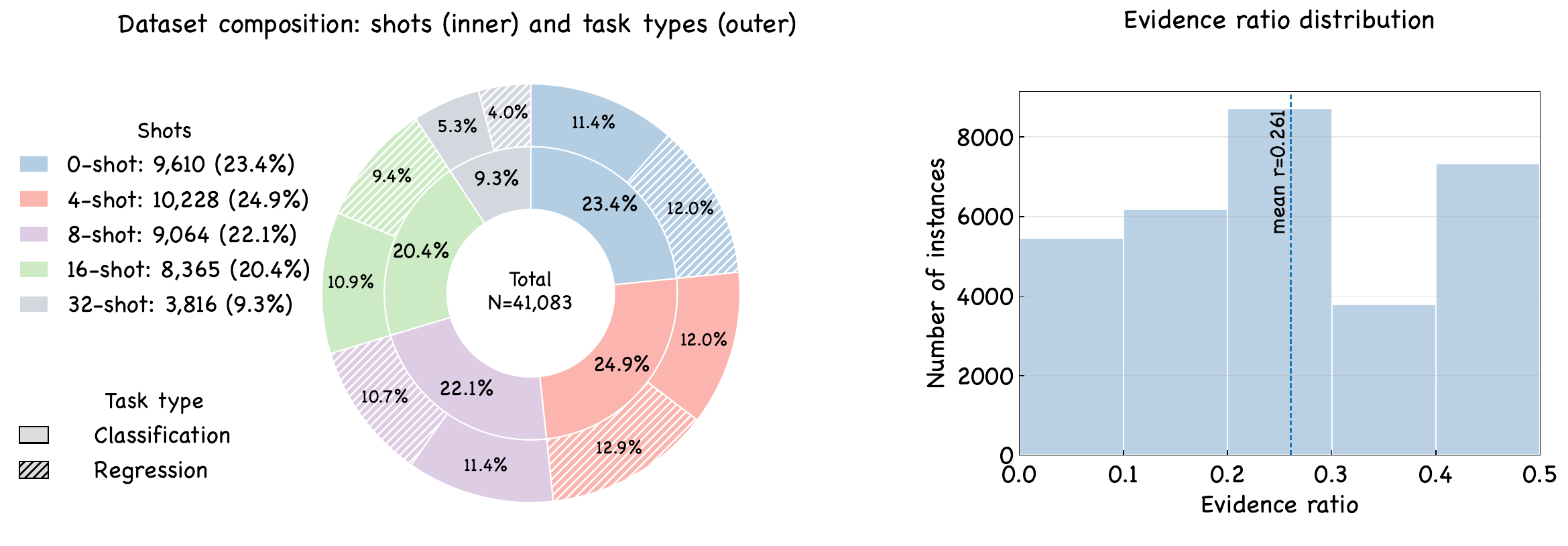}
    \captionof{figure}{Data distribution of the SFT dataset. \textbf{Left:} dataset
    composition across the five shot budgets, with each slice further decomposed
    into classification and regression trajectories. \textbf{Right:}
    distribution of the evidence ratio for $K>0$.}
    \label{fig:dataset}
\end{minipage}
\end{center}
\end{@twocolumnfalse}%
}]

\section{Implementation Details.}
\subsection{Details of Feature Filtering}
\label{sec:mi_filter}
We consider a labeled tabular dataset $\mathcal{D}=\{(x_i,y_i)\}_{i=1}^{N}$ with $N$ instances and $d$ feature columns. We first discard identifier-like columns, such as user IDs or email addresses, since their near-unique values can spuriously yield high mutual information with the target and dominate MI-based ranking, despite being uninformative for prediction.
After this removal, we retain $d'$ candidate feature columns. 
We denote the $j$-th remaining feature column as a random variable $X_j$ and the target column as $Y$. 
For each candidate feature $X_j$, we compute its MI score with the target as:
\begin{equation}
s_j \;=\; I(X_j;Y) \;=\; \sum_{u}\sum_{v} p(u,v)\,\log\frac{p(u,v)}{p(u)\,p(v)},
\end{equation}
where $u$ and $v$ enumerate possible values of $X_j$ and $Y$, respectively.

We then rank features by MI scores in descending order, obtaining $s_{(1)} \ge \cdots \ge s_{(d')}$. 
We select the smallest prefix whose cumulative MI reaches a fixed fraction $\rho$ of the total MI mass:
\begin{equation}
m^\star \;=\; \min\left\{m \,\middle|\, \sum_{t=1}^{m} s_{(t)} \ge \rho \sum_{t=1}^{d'} s_{(t)}\right\},
\end{equation}
where $\rho=0.9$. Finally, we keep the top-$m^\star$ features together with the target column to form the filtered table. 
This procedure removes weakly related fields and improves the signal-to-noise ratio, while maintaining diverse dependency patterns and moderate residual noise that better reflects real-world tabular prediction scenarios.

\subsection{Distribution of \sftname}
\label{sec:distru}
We summarize the data distribution of our SFT dataset in Figure~\ref{fig:dataset}. After strict trajectory filtering, \sftname contains $41{,}083$ select-then-predict trajectories. As shown in the left panel, each instance is constructed under a shot budget $K\in\{0,4,8,16,32\}$ and belongs to either a classification or a regression task. Across the five shot budgets, the dataset contains $9{,}610$, $10{,}228$, $9{,}064$, $8{,}365$, and $3{,}816$ instances, corresponding to $23.4\%$, $24.9\%$, $22.1\%$, $20.4\%$, and $9.3\%$. The task types are approximately balanced overall, and both types are present under every shot budget. For $K>0$, we further report the evidence ratio $r$, defined as the fraction of evidential rows among the sampled in-context candidates. As shown in the right panel of Figure~\ref{fig:dataset}, $r$ ranges from $0$ to $0.5$ with an average value of $\bar{r}=0.261$. The histogram exhibits a broad coverage of evidence densities, including both sparse-evidence cases and relatively dense-evidence cases, which improves distributional diversity and better reflects realistic tabular prediction scenarios where useful context can be sparse and unevenly distributed.

\begin{table}[!t]
\centering
\setlength{\tabcolsep}{4pt}
\renewcommand{\arraystretch}{1.05}
\begin{tabular}{@{}l|c@{}}
\toprule
\textbf{Hyperparameters} & \textbf{Value} \\
\midrule
Epochs & 4.0 \\
Micro batch size & 4 \\
Gradient accumulation & 4 \\
Learning rate & $1 \times 10^{-4}$ \\
Warmup ratio & 0.1 \\
LR scheduler & Cosine \\
GPUs & $8 \times \texttt{H200}$ \\
Training time (h) & 14.11 \\
\bottomrule
\end{tabular}
\caption{Core SFT hyperparameters for cold-start.}
\label{tab:sft_hparams_compact}
\end{table}

\subsection{Reinforcement Learning Dataset Construction}
\label{sec:rldata}
We build the reinforcement learning training set \rlname by filtering instances that can destabilize optimization or provide limited learning signal. We first discard examples for which the teacher model fails to produce an acceptable prediction within five attempts during SFT trajectory synthesis, as these cases often indicate insufficient supporting evidence, excessive difficulty, or an ill-posed target. We further remove examples that Qwen3-8B~\cite{qwen3} solves correctly with a single forward evaluation. Since Qwen3-8B shares the same pretraining knowledge as our backbone, such instances are typically near-trivial and yield limited benefit from reinforcement learning. After filtering, \rlname concentrates on medium difficulty instances, which improves training stability and makes the learning signal more informative.

\begin{table}[!t]
\centering
\setlength{\tabcolsep}{5pt}
\renewcommand{\arraystretch}{1.05}
\begin{tabular}{@{}l|c@{}}
\toprule
\textbf{Hyperparameters} & \textbf{Value} \\
\midrule
Epochs & 2.0 \\
Train batch size & 256 \\
Rollout sampling num & 8 \\
Rollout temperature & 1.0 \\
KL coefficient & 0.001 \\
Learning rate & $1 \times 10^{-6}$ \\
Max prompt length & 4096 \\
Max response length & 5120 \\
Clip ratio $\epsilon_{\mathrm{low}}$ & 0.2 \\
Clip ratio $\epsilon_{\mathrm{high}}$ & 0.28 \\
$\lambda_f$ & 0.1 \\
$\lambda_e$ & 0.2 \\
$\lambda_c$ & 0.7 \\
$\gamma$ & 1.0 \\
$\beta$ & 0.99 \\
$\eta_{\min}$ & 0.8 \\
$\eta_{\max}$ & 1.0 \\
GPUs & $8 \times \texttt{H200}$ \\
Training time (h) & 25.08 \\
\bottomrule
\end{tabular}
\caption{Core RL hyperparameters for TAB-GRPO.}
\label{tab:rl_hparams_compact}
\end{table}

% \begin{table}[htbp]
% \centering
% \vspace{-0.2cm}
% \caption{Core RL hyperparameters for TAB-GRPO.}
% \setlength{\tabcolsep}{3pt}
% \begin{tabular}{c c | c c}
% \toprule
% \textbf{Hyper-parameters} & \textbf{Value} &
% \textbf{Hyper-parameters} & \textbf{Value} \\
% \midrule
% Epochs & 2.0 & $\lambda_f$ & 0.1 \\
% Train batch size & 256 & $\lambda_e$ & 0.2 \\
% Rollout Sampling Num & 8 & $\lambda_c$ & 0.7 \\
% Rollout temperature & 1.0 & $\gamma$ & 1.0 \\
% KL coefficient & 0.001 & $\beta$ & 0.99 \\
% Learning rate & $1 \times 10^{-6}$ & $\eta_{min}$ & 0.8 \\
% Max prompt length & 4096 & $\eta_{max}$ & 1.0 \\
% Max response length & 5120 & GPUs & $8 \times \texttt{H200}$ \\
% Clip ratio (low) $\epsilon_{\text{low}}$ & 0.2 & Training time (h) & 25.08 \\
% Clip ratio (high) $\epsilon_{\text{high}}$ & 0.28 & & \\
% \bottomrule
% \end{tabular}
% \vspace{-0.2cm}
% \label{tab:rl_hparams_compact}
% \end{table}

\subsection{Parameters Setup}
\label{sec:hyper}
We report the key hyperparameters used in our two-stage training pipeline, and present the essential settings for SFT and RL training with TAB-GRPO in Tables~\ref{tab:sft_hparams_compact} and~\ref{tab:rl_hparams_compact}, respectively. In particular, we set $\epsilon_{\mathrm{high}}$ to 0.28 following~\cite{dapo} to encourage exploration, and set the EMA decay factor $\beta$ to 0.99 following common practice. Omitted options follow the default values of the training framework.

\subsection{Prompt Template}
In this section, we describe the prompt templates used in our experiments. Figure~\ref{fig:tabsieve_prompt} presents the prompt for \name. It first asks the model to analyze the table structure and the relationships among feature columns, and then to select evidential rows based on this understanding before making the final prediction. The model is finally required to output the answer in the \texttt{\textbackslash boxed\{\}} format. Figure~\ref{fig:general_prompt} shows the prompt used for the general LLM baselines. Compared with the prompt of \name, we only remove the instructions related to evidence selection, so as to minimize prompt-induced bias. For the tabular LLM baselines, we follow the prompt templates provided in their official implementations for inference.

\begin{figure*}[h]
\centering

\begin{promptbox}{Prompt Template}
\textbf{Role}

You are an expert in Data Mining and Logical Reasoning. Your core expertise lies in deeply comprehending the intrinsic logic of tabular data, analyzing the structural and feature-target dependencies, identifying samples similar to the target, and utilizing these insights to predict the `[MISSING]` value.

\textbf{Task Description}

Given a sampled sub-table and its metadata:

\begin{itemize}
  \item The last cell of the final row is marked as `[MISSING]`, serving as the prediction target.
  \item You must conduct a step-by-step analysis within `<think>` tags, adhering to the `Reasoning Requirements`, to derive a \textbf{specific} predicted value.
\end{itemize}

\textbf{Reasoning Requirements}
\begin{itemize}
  \item Comprehend the table structure and the semantics of feature columns.
  \item Analyze potential relationships between features and the target (e.g., strong correlations, causality).
  \item Identify samples similar to the target. During the reasoning process, enclose the indices of these selected samples within `<select>` tags (e.g., `<select> 2, 4, 7 </select>`).
  \item Synthesize the analysis to model the logic connecting features to the target and predict the `[MISSING]` value.
  \item Construct a complex and comprehensive logical derivation from features to the target, simulating a predictive model.
  \item Reject lazy reasoning: Deriving the result solely by calculating the mean, median, or mode of similar samples is strictly prohibited.
\end{itemize}

\textbf{Output Format}

Output ONLY the final answer, excluding any additional text, and put your final answer within `$\backslash$boxed\{\}`.

\textbf{Metadata of the table named} 

\texttt{\{table\_name\}}

\texttt{\{meta\_data\}}

\textbf{Sampled Sub-table}

\texttt{\{sub\_table\}}

\textbf{Candidate Values}

\texttt{\{range\_of\_target\}}

Please reason step by step, output ONLY the final answer, put your final answer within $\backslash$boxed\{\}.
\end{promptbox}
\caption{Training and inference prompt templates of the \name.}
\label{fig:tabsieve_prompt}
\end{figure*}

\begin{figure*}[h]
\centering

\begin{promptbox}{Prompt Template}
\textbf{Role}

You are an expert in Data Mining and Logical Reasoning. Your core expertise lies in deeply comprehending the intrinsic logic of tabular data, analyzing the structural and feature-target dependencies, and predicting the `[MISSING]` value.

\textbf{Task Description}

Given a sampled sub-table and its metadata:

\begin{itemize}
  \item The last cell of the final row is marked as `[MISSING]`, serving as the prediction target.
  \item You must conduct a step-by-step analysis within `<think>` tags, adhering to the `Reasoning Requirements`, to derive a \textbf{specific} predicted value.
\end{itemize}

\textbf{Reasoning Requirements}
\begin{itemize}
  \item Comprehend the table structure and the semantics of feature columns.
  \item Analyze potential relationships between features and the target (e.g., strong correlations, causality).
  \item Synthesize the analysis to model the logic connecting features to the target and predict the `[MISSING]` value.
  \item Construct a complex and comprehensive logical derivation from features to the target, simulating a predictive model.
  \item Reject lazy reasoning: Deriving the result solely by calculating the mean, median, or mode of similar samples is strictly prohibited.
\end{itemize}

\textbf{Output Format}

Output ONLY the final answer, excluding any additional text, and put your final answer within `$\backslash$boxed\{\}`.

\textbf{Metadata of the table named} 

\texttt{\{table\_name\}}

\texttt{\{meta\_data\}}

\textbf{Sampled Sub-table}

\texttt{\{sub\_table\}}

\textbf{Candidate Values}

\texttt{\{range\_of\_target\}}

Please reason step by step, output ONLY the final answer, put your final answer within $\backslash$boxed\{\}.
\end{promptbox}
\caption{Inference prompt templates of the general LLMs.}
\label{fig:general_prompt}
\end{figure*}

% \begin{figure}[hbtp]
%   \centering
%   \includegraphics[width=0.9\linewidth]{figures/zero-case-cls.pdf}
%   \vspace{-0.3cm}
% % \caption{Distribution of SFT dataset.}
% \caption{Zero-shot case in classification tasks.}
%     % \vspace{-0.3cm}
%   \label{fig:zero-case-cls}
% \end{figure}

% \begin{figure}[h]
%   \centering
%   \includegraphics[width=0.85\linewidth,keepaspectratio]{figures/zero-case-reg.pdf}
%   \vspace{-0.3cm}
% % \caption{Distribution of SFT dataset.}
% \caption{Zero-shot case in regression tasks.}
%     % \vspace{-0.3cm}
%   \label{fig:zero-case-reg}
% \end{figure}

% \begin{figure}[h]
%   \centering
%   \includegraphics[width=0.81\linewidth]{figures/few-shot-cls.pdf}
%   \vspace{-0.3cm}
% % \caption{Distribution of SFT dataset.}
% \caption{Few-shot case in classification tasks.}
%     % \vspace{-0.3cm}
%   \label{fig:few-shot-cls}
% \end{figure}

\begin{table}[!t]
\centering
\begingroup
\footnotesize
\setlength{\tabcolsep}{4.5pt}
\begin{tabular}{lcccc}
\toprule
\textbf{Evidence}  &  \textbf{4} & \textbf{8} & \textbf{16} & \textbf{32} \\
\midrule

$\lvert \mathcal{E}^{\mathrm{inf}} \rvert$ &  4592 & 8594 & 17791 & 34313  \\
$\lvert \mathcal{E}^{\mathrm{emb}} \rvert$ & 5681 & 10327 & 23717 & 46235 \\

$\lvert \mathcal{E}^{\mathrm{emb}} \cap \mathcal{E}_i^{\mathrm{inf}}\rvert$ & 3551 & 6816 & 14968 & 28567 \\
\midrule
$\frac{\lvert \mathcal{E}^{\mathrm{emb}} \cap \mathcal{E}^{\mathrm{inf}}\rvert}{\lvert \mathcal{E}^{\mathrm{inf}} \rvert}$ & 77.33\% & 79.31\% & 84.13\% & 83.25\% \\

\bottomrule
\end{tabular}
\endgroup
\caption{Overlap between embedding-based evidence and influence-based evidence under different shot budgets.}
\label{tab:emb_inf_overlap}
\end{table}

\begin{table}[!t]
\centering
\begingroup
\footnotesize
\setlength{\tabcolsep}{4.5pt}
\begin{tabular}{lccccc}
\toprule
\textbf{Evidence}  & \textbf{0} & \textbf{4} & \textbf{8} & \textbf{16} & \textbf{32} \\
\midrule

Qwen3-8B   & 48.18 & 58.77 & 60.91 & 63.01 & 65.67  \\
\midrule
Influence-based  &  54.41 & 65.23 & 69.06 & 70.72 & 74.42  \\
Embedding-based  &  54.63 & 65.77 & 69.32 & 70.98 & 74.33  \\

\bottomrule
\end{tabular}
\endgroup
\caption{Classification results of different evidence construction strategies. We report accuracy (↑) averaged over all classification datasets under all shot settings.}
\label{tab:ablation_cls_evidence}
\end{table}

\begin{table}[!t]
\centering
\begingroup
\footnotesize
\setlength{\tabcolsep}{4.5pt}
\begin{tabular}{lccccc}
\toprule
\textbf{Evidence}  & \textbf{0} & \textbf{4} & \textbf{8} & \textbf{16} & \textbf{32} \\
\midrule

Qwen3-8B & 0.325 & 0.268 & 0.239 & 0.235 & 0.222 \\
\midrule
Influence-based &  0.262 & 0.184 & 0.171 & 0.149 & 0.132  \\
Embedding-based &  0.266 & 0.192 & 0.169 & 0.158 & 0.139 \\

\bottomrule
\end{tabular}
\endgroup
\caption{Regression results of different evidence construction strategies. We report NMAE (↓) averaged over all regression datasets under all shot settings.}
\label{tab:ablation_reg_evidence}
\end{table}

\section{Additional Analysis}
\subsection{Comparison with Influence-Based Evidence}
\label{app:influence_evidence}
In the main experiments, we use embedding-based evidence as the supervision signal for explicit row selection. 
Although Section~\ref{sec:diag} has shown that embedding-based evidence can improve prediction, one may ask whether a more direct, performance-based criterion can provide better evidence labels. 
To further validate the reasonableness of embedding-based evidence, we compare it with an influence-based evidence construction strategy.

For each training instance $i$, let $\mathcal{E}_i^{\mathrm{emb}}$ denote the evidence set selected by the embedding-based construction, and let $k_i = |\mathcal{E}_i^{\mathrm{emb}}|$ be the corresponding evidence budget.
We construct an influence-based evidence set $\mathcal{E}_i^{\mathrm{inf}}$ using a leave-one-out criterion.
Specifically, we first run Qwen3-8B on the original context and compute the Avg@5 performance.
We then remove each candidate row $x_{i,j}$ from the context and rerun the same Avg@5 evaluation.
The influence of $x_{i,j}$ is defined by the performance degradation caused by removing it.
We construct $\mathcal{E}_i^{\mathrm{inf}}$ by selecting up to $k_i$ rows with the largest positive influence scores.
Note that not every row removal leads to a performance drop.
Therefore, for some instances, the number of positively influential rows is smaller than $k_i$.

Table~\ref{tab:emb_inf_overlap} shows that $\mathcal{E}^{\mathrm{emb}}$ covers a substantial portion of the influential rows. This suggests that $\mathcal{E}^{\mathrm{emb}}$ already captures many rows that are important for prediction, while much cheaper to construct. Moreover, as shown in Tables~\ref{tab:ablation_cls_evidence} and~\ref{tab:ablation_reg_evidence}, $\mathcal{E}^{\mathrm{inf}}$ achieves comparable performance to $\mathcal{E}^{\mathrm{emb}}$, but does not yield a consistent advantage. 
Since constructing $\mathcal{E}^{\mathrm{inf}}$ requires repeated leave-one-out inference, we use embedding-based evidence in the main experiments as an effective and efficient supervision signal for evidence selection.

\subsection{More Ablation Studies}

\begin{table}[!t]
\centering
\begingroup
\footnotesize
\setlength{\tabcolsep}{5pt}
\begin{tabular}{lcc}
\toprule
\textbf{Model} & \textbf{Acc. $\uparrow$} & \textbf{NMAE $\downarrow$} \\
\midrule
Qwen2.5-72B w. Std. Prompt & 65.11 & 0.217 \\
Qwen2.5-72B w. TabSieve Prompt & 65.92 & 0.209 \\
Qwen3-8B w. Std. Prompt & 59.31 & 0.258 \\
Qwen3-8B w. TabSieve Prompt & 59.79 & 0.251 \\
\midrule
\name (Ours) & \textbf{67.01} & \textbf{0.185} \\
\bottomrule
\end{tabular}
\endgroup
\caption{Prompt ablation. We report average classification accuracy and regression NMAE over all shot settings.}
\label{tab:prompt_ablation}
\end{table}
\paragraph{Prompt Design.} Table~\ref{tab:prompt_ablation} compares the standard prompt with the TabSieve prompt on general LLMs. 
Using the TabSieve prompt improves the performance, indicating that explicit instructions for evidence selection are beneficial for tabular prediction. 
However, prompt engineering alone still falls behind \name, suggesting that jointly optimizing evidence selection and prediction is still necessary for achieving more robust cross-table generalization.

\begin{table}[!t]
\centering
\begingroup
\footnotesize
\setlength{\tabcolsep}{5pt}
\begin{tabular}{lccc}
\toprule
\textbf{Hyperparameter} & \textbf{Value} & \textbf{Acc. $\uparrow$} & \textbf{NMAE $\downarrow$} \\
\midrule
$G$ & 4  & 65.97 & 0.209 \\
 & 8  & \textbf{67.01} & 0.185 \\
& 12 & 66.82 & \textbf{0.179} \\
\midrule
$\eta_{\min}$ & 0.7 & 66.57 & 0.197 \\
 & 0.8 & \textbf{67.01} & \textbf{0.185} \\
& 0.9 & 66.14 & 0.192 \\
\bottomrule
\end{tabular}
\endgroup
\caption{Sensitivity analysis of key RL hyperparameters.}
\label{tab:parameter}
\end{table}

\paragraph{Effects of group size $G$ and lower bound $\eta_{min}$ in TAB-GRPO.}
Table~\ref{tab:parameter} presents the sensitivity analysis of two key hyperparameters in RL training. 
Increasing the group size $G$ generally leads to better performance, since a larger group provides more diverse rollouts and more reliable group-relative advantage estimation. 
However, it also increases the computational cost of each update. 
The lower bound $\eta_{\min}$ controls the minimum scale of regression advantages in the task-advantage balancing mechanism. 
When $\eta_{\min}$ is too small, regression samples may be overly suppressed in the early stage, reducing learning efficiency. 
When $\eta_{\min}$ is too large, the balancing effect becomes weaker, which limits its ability to mitigate the optimization imbalance between classification and regression.

\begin{table}[!t]
\centering
\begingroup
\footnotesize
\setlength{\tabcolsep}{5pt}
\begin{tabular}{lcc}
\toprule
\textbf{Variant} & \textbf{Acc. $\uparrow$} & \textbf{NMAE $\downarrow$} \\
\midrule
w/o EMA Smoothing & 65.12 & 0.207 \\
w/o $\eta_{\min}$ & 66.04 & 0.202 \\
\midrule
\name & \textbf{67.01} & \textbf{0.185} \\
\bottomrule
\end{tabular}
\endgroup
\caption{Component-level ablation of the task-advantage balancing mechanism in TAB-GRPO. We report average classification accuracy and regression NMAE over all shot settings.}
\label{tab:tabgrpo_component}
\end{table}

\paragraph{Component-level ablation of TAB-GRPO.}
The task-advantage balancing mechanism of TAB-GRPO combines two ingredients: the EMA of Eq.~(\ref{eq:ema_adv}), which tracks the task-wise advantage scale with decay $\beta$, and the lower bound $\eta_{\min}$ on the rescaling factor of Eq.~(\ref{eq:task_scaler}).
To isolate their individual contributions, we conduct component-level ablations in Table~\ref{tab:tabgrpo_component}.
The \emph{w/o EMA Smoothing} variant sets $\beta=0$, which reduces $m_t^{\tau}$ to the current-step statistics and is therefore more sensitive to local fluctuations.
The \emph{w/o $\eta_{\min}$} variant removes the lower-bound protection, allowing the regression advantage to become too small at some training steps and weakening its effective learning signal.
Both variants underperform the full method, supporting the contributions of the EMA estimate and the bounded rescaling factor.
The sensitivity analysis in Table~\ref{tab:parameter} also provides additional evidence for the bound, since both an overly small and an overly large $\eta_{\min}$ degrade performance relative to the chosen value.

\begin{table}[!t]
\centering
\begingroup
\footnotesize
\setlength{\tabcolsep}{3.5pt}
\begin{tabular}{lccccc}
\toprule
\textbf{Model} & \textbf{0} & \textbf{4} & \textbf{8} & \textbf{16} & \textbf{32} \\
\midrule
DeepSeek-R1-70B & \underline{41.06} & \underline{52.21} & \underline{54.75} & \underline{53.12} & \underline{59.93} \\
GPT-OSS-20B & 40.75 & 48.99 & 50.15 & 52.48 & 53.87 \\
Qwen3-8B & 26.76 & 34.62 & 43.22 & 47.74 & 50.92 \\
\midrule
\name & \textbf{43.19} & \textbf{54.10} & \textbf{56.06} & \textbf{55.92} & \textbf{60.31} \\
\bottomrule
\end{tabular}
\endgroup
\caption{Macro-F1 ($\uparrow$) averaged over all classification datasets under zero-shot and few-shot settings. The best result in each column is in bold and the second best is underlined.}
\label{tab:macro_f1}
\end{table}

\subsection{Macro-F1 on Classification Tasks}
\label{app:macro_f1}
Accuracy can be dominated by majority classes on imbalanced tables, so it does not fully characterize classification quality.
We therefore additionally report Macro-F1, which weights every class equally regardless of its frequency.
Table~\ref{tab:macro_f1} compares \name with the strongest general LLM baselines and with the Qwen3-8B backbone under all shot budgets.
\name achieves the best Macro-F1 under every shot budget.
The margin over the backbone is especially large in the low-shot regime, where Qwen3-8B tends to collapse onto the majority class, while \name still recovers minority classes by grounding its prediction on explicitly selected evidence rows.
These results confirm that \name has stronger classification performance under a class-balanced evaluation, and that its accuracy gains in Table~\ref{tab:main_rst} are not an artifact of majority-class bias.

\begin{table}[!t]
\centering
\begingroup
\footnotesize
\setlength{\tabcolsep}{5pt}
\begin{tabular}{lcc}
\toprule
\textbf{Variant} & \textbf{Acc. $\uparrow$} & \textbf{NMAE $\downarrow$} \\
\midrule
Qwen3-8B & \textbf{59.31} & \textbf{0.258} \\
Qwen3-8B w/o Selected Rows & 43.57 & 0.336 \\
\bottomrule
\end{tabular}
\endgroup
\caption{Utility of the rows selected by \name. We remove the selected rows from the context and re-evaluate Qwen3-8B on the remaining rows. Results are averaged over all shot settings.}
\label{tab:row_utility}
\end{table}

\subsection{Utility of the Selected Rows}
\label{app:row_utility}
We further evaluate the utility of the rows selected by \name.
Specifically, for each query we take the evidence rows that \name selects, remove these rows from the in-context support set, and evaluate Qwen3-8B on the remaining context.
As shown in Table~\ref{tab:row_utility}, removing the selected rows results in $26.5\%$ lower accuracy and $30.2\%$ higher NMAE, directly showing that the selected rows contain useful predictive information.

\subsection{More Case Studies}
\label{sec:more_case}
\begin{figure*}[t]
\centering
\begin{casebox}{1:}

\noindent
\begin{minipage}{0.35\linewidth}
    \includegraphics[width=\linewidth]{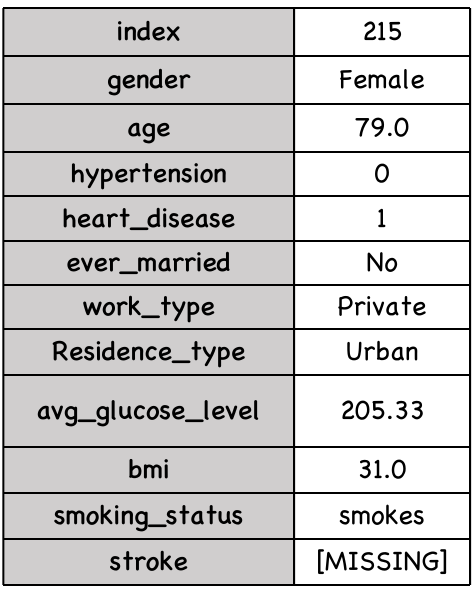}
\end{minipage}
\hfill
\begin{minipage}{0.62\linewidth}
    {\textbf{Question}}\par\vspace{0.7em}

    {\textbf{Metadata of the table named stroke-prediction-dataset.}}

    {The column 'age' represents 'Age of the individual in years' and is a 'numerical' variable.} \par\vspace{0.7em}

    {...} \par\vspace{0.7em}

    {\textbf{Candidate Values}}

    {The possible values are: '1', '0'.} \par\vspace{0.7em}
    {Please reason step by step, output ONLY the final answer, put your final answer within `$\backslash$boxed\{\}`.} \par\vspace{1em}
    {\textbf{Ground Truth: } $1$}
\end{minipage}

\par\medskip

\noindent
\begin{minipage}{\linewidth}
\small

\par\smallskip
\hrule
\par\smallskip

{\bfseries TabSieve Output.}

\texttt{<think>}

Okay, let's tackle this problem step by step. The task is to predict the missing stroke value for the target row...
\par\smallskip
\par\smallskip
First, looking at the table structure. The dataset is called stroke-prediction-dataset. The target column is 'stroke', which is categorical with possible values 0 or 1...
\par\smallskip
\par\smallskip
Now, considering the relationships between features and the target. Based on general knowledge, stroke risk factors typically include hypertension (high blood pressure), heart disease, smoking... For example, hypertension is a major risk factor for stroke. Similarly, having a heart disease could increase the risk...

\par\smallskip
\par\smallskip

\textcolor[HTML]{8C564B}{<select></select>}

\par\smallskip
\par\smallskip

\textcolor[HTML]{2CA02C}{Stroke risk factors include age (older age increases risk), heart disease (yes), smoking, high glucose (diabetes), BMI (obesity), etc...However, heart disease is present. Also, smoking is a risk factor. High glucose could indicate diabetes, which is a risk factor. Age 79 is very high risk...}

\par\smallskip
\par\smallskip

Given that heart disease is a strong predictor of stroke, and she has it, and she smokes, the risk is elevated. High glucose further increases this risk. \textcolor[HTML]{17BECF}{Therefore, despite no hypertension, the overall risk is high, and the most probable outcome is stroke=1.}

\texttt{</think>}
\par\medskip
\noindent $\boxed{1}$

\end{minipage}
\end{casebox}
\caption{Zero-shot case in classification tasks. \name first analyzes the semantic meaning of the patient attributes and then predicts the label based on medical knowledge about stroke risk.}
\label{fig:zero-case-cls}
\end{figure*}

\begin{figure*}[t]
\centering
\begin{casebox}{2:}

\noindent
\begin{minipage}{0.35\linewidth}
    \includegraphics[width=\linewidth]{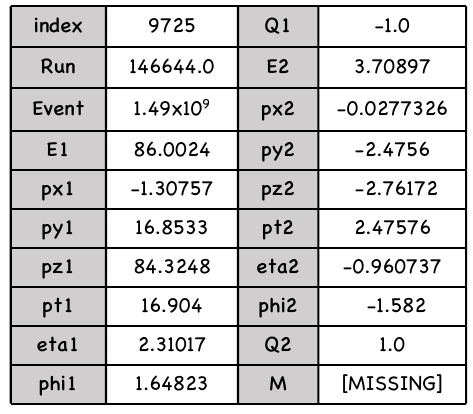}
\end{minipage}
\hfill
\begin{minipage}{0.62\linewidth}
    {\textbf{Question}}\par\vspace{0.7em}

    {\textbf{Metadata of the table named cern-electron-collision-data.}}

    {The column 'E1' represents 'Energy of electron 1' and is a 'numerical' variable. The column 'px1 ' represents 'Momentum of electron 1 in the x direction'...} \par\vspace{0.7em}

    {...} \par\vspace{0.7em}

    {\textbf{Candidate Values}}

    {The value of M ranges from 2.00008 to 109.999 (inclusive).} \par\vspace{0.7em}
    {Please reason step by step, output ONLY the final answer, put your final answer within `$\backslash$boxed\{\}`.} \par\vspace{1em}
    {\textbf{Ground Truth: } $34.4543$}
\end{minipage}

\par\medskip

\noindent
\begin{minipage}{\linewidth}
\small

\par\smallskip
\hrule
\par\smallskip

{\bfseries TabSieve Output.}

\texttt{<think>}

Okay, let's tackle this problem step by step.First, let's look at the metadata. The table is about CERN electron collision data. The target column is 'M', which is the invariant mass of the two-electron system...
\par\smallskip
\par\smallskip
Now, the target is M, the invariant mass. In physics, the invariant mass of a system is calculated from the total energy and momentum. For two particles, the formula is $M = sqrt((E1 + E2)^2 - (p\_x1 + p\_x2)^2 - (p\_y1 + p\_y2)^2 - (p\_z1 + p\_z2)^2)$. So M should be directly determined by the sum of the energies and the sum of the momenta components...

\par\smallskip
\par\smallskip

\textcolor[HTML]{8C564B}{Since there are no other samples available for comparison, no samples can be selected. <select></select>}

\par\smallskip
\par\smallskip

...So the answer should be the computed value of M using the formula. Let's compute it for the given row...
\textcolor[HTML]{2CA02C}{First, sum of energies: E1 + E2 = 86.0024 + 3.70897 = 89.71137...
Now, compute the sum of squared momenta: $(px1+px2)^2 = (-1.3353026)^2 \approx 1.7830$...
Total sum of squared momenta: $1.7830 + 206.72 + 6652.54 \approx 6861.043$....
Then $M^2 = 8048.140 - 6861.046 = 1187.097$ and
$M = sqrt(1187.097) \approx 34.454$}

\par\smallskip
\par\smallskip

This value falls within the expected range of 2.00008 to 109.999, confirming the calculation is consistent. \textcolor[HTML]{17BECF}{Therefore, the missing value is approximately 34.454.}

\texttt{</think>}
\par\medskip
\noindent $\boxed{34.454}$

\end{minipage}
\end{casebox}
\caption{Zero-shot case in regression tasks. \name recognizes that the target is the invariant mass of a two-electron collision system, and then derives the final prediction by applying the physics formula.}
\label{fig:zero-case-reg}
\end{figure*}

\begin{figure*}[t]
\centering
\begin{casebox}{3:}

\noindent
\begin{minipage}{\linewidth}
    \centering
    \includegraphics[width=1.0\linewidth]{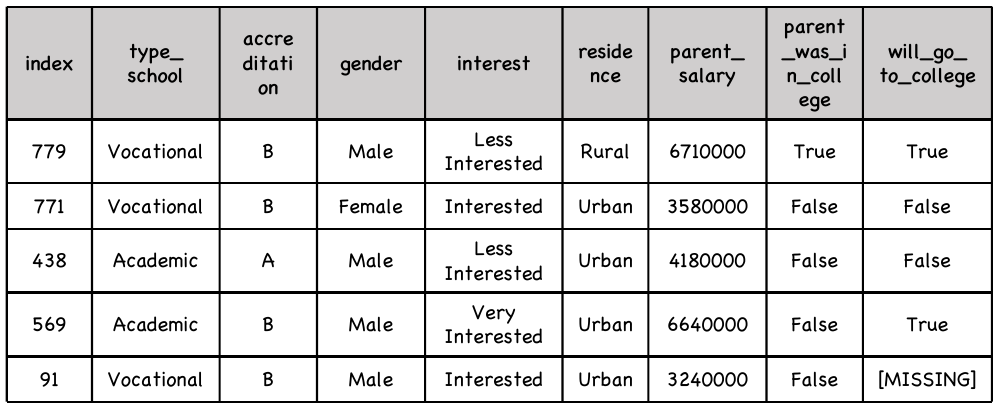}
\end{minipage}

\vspace{0.9em}

\noindent
\begin{minipage}{\linewidth}
    {\textbf{Question}}\par\vspace{0.7em}
    
    {\textbf{Metadata of the table named go-to-college-dataset.}}\par\vspace{0.4em}

    The column 'type\_school' represents 'The type of school the student attends, either Academic or Vocational' and is a 'categorical' variable...\par\vspace{0.7em}

    {...}\par\vspace{0.7em}

    {\textbf{Candidate Values}}\par\vspace{0.4em}

    The possible values are: 'True', 'False'.\par\vspace{0.7em}

    Please reason step by step, output ONLY the final answer, put your final answer within `$\backslash$boxed\{\}`.\par\vspace{1em}

    {\textbf{Ground Truth:} $False$}
\end{minipage}

\par\medskip
\hrule
\par\smallskip

\noindent
\begin{minipage}{\linewidth}
\small

{\bfseries TabSieve Output.}

\texttt{<think>}

Okay, let's tackle this problem step by step. The task is to predict the missing information in the 'will\_go\_to\_college' column for the target sample...Let's think about general knowledge first. Typically, the type of school...urban vs rural could have different access to college resources. Parent's salary might indicate financial ability to afford college. If the parent was in college, maybe that's a positive indicator for the student to go too...

\par\smallskip
\par\smallskip

Line 771: Vocational, B... So same school type (Vocational), same school accreditation (B)... But maybe gender isn't a strong factor here. \textcolor[HTML]{8C564B}{So line 771 is very similar}...
Line 569: Academic, B,... So same gender (Male), same school accreditation (B), similar interest (Very Interested vs Interested), same residence (Urban), parent not in college. Parent salary higher. \textcolor[HTML]{8C564B}{So maybe this is the second similar...}

\par\smallskip
\par\smallskip

\textcolor[HTML]{8C564B}{<select> 771, 569 </select>}

\par\smallskip
\par\smallskip

Line 771 has the same school type... The only difference is gender, but gender might not be a strong predictor. For example, line 438 is male...So gender might not be a major factor here...\textcolor[HTML]{2CA02C}{So maybe for Vocational, if parent not in college, they don't go to college. So target would be False.}
But line 569 is Academic, B...but went True...\textcolor[HTML]{2CA02C}{So school type and salary matter.}

\par\smallskip
\par\smallskip
\textcolor[HTML]{17BECF}{So for the target, which is Vocational, B, parent not in college. So maybe target is False.}

\texttt{</think>}

\par\medskip
\noindent $\boxed{False}$

\end{minipage}

\end{casebox}
\caption{Few-shot case. \name follows a select-then-predict procedure. It first analyzes the table semantics and the influence of features on the target, then selects the evidence rows, and finally performs prediction mainly based on these selected rows.}
\label{fig:few-shot-cls}
\end{figure*}

In this section, we present additional examples to illustrate the reasoning patterns of \name under different inference settings. Figure~\ref{fig:zero-case-cls} and Figure~\ref{fig:zero-case-reg} show \name in the zero-shot regime. Since no examples are available for in-context learning, the model relies entirely on general knowledge to model relationships among the features and predicts the missing value based on the inferred mapping. Figure~\ref{fig:few-shot-cls} presents a few-shot example, where the model first infers and constructs feature relations from the context, then selects a small set of evidential rows, and finally uses the selected evidence to produce the final prediction.

\end{document}